\documentclass[mnsc,nonblindrev]{informs3} 

\usepackage[dvipsnames]{xcolor}
\usepackage{float}

\usepackage{algorithm}
\usepackage{algpseudocode}
\usepackage{bm}
\usepackage{booktabs}
\usepackage{hyperref}
\usepackage{graphicx}
\usepackage{enumitem}
\usepackage{multirow}
\usepackage{comment}
\usepackage{pifont}
\usepackage{soul}

\OneAndAHalfSpacedXI

\usepackage{natbib}
 \bibpunct[, ]{(}{)}{,}{a}{}{,}%
 \def\bibfont{\small}%
\TheoremsNumberedThrough     
\ECRepeatTheorems

\EquationsNumberedThrough    

\newcommand{\bx}{\mbox{\bf x}}

\newcommand{\bX}{\mbox{\bf X}}

\newcommand{\cmark}{\ding{51}} 
\newcommand{\xmark}{\ding{55}} 
\MANUSCRIPTNO{MS-0001-1922.65}

\begin{document}



\RUNTITLE{Tis a Butter Place}

\TITLE{Deep Generative Demand Learning for Newsvendor and Pricing}

\ARTICLEAUTHORS{%
\AUTHOR{Shijin Gong}
\AFF{School of Management, University of Science and Technology of China, Hefei, China, \EMAIL{shijin49@mail.ustc.edu.cn}} 
\AUTHOR{Huihang Liu}
\AFF{International Institute of Finance, School of Management, University of Science and Technology of China, Hefei, China, \EMAIL{huihang@mail.ustc.edu.cn}}
\AUTHOR{Xinyu Zhang}
\AFF{Academy of Mathematics and Systems Science, Chinese Academy of Sciences, Beijing, China, \EMAIL{xinyu@amss.ac.cn}}
} 

\ABSTRACT{%
We consider data-driven inventory and pricing decisions in the feature-based newsvendor problem, where demand is influenced by both price and contextual features and is modeled without any structural assumptions. The unknown demand distribution results in a challenging conditional stochastic optimization problem, further complicated by decision-dependent uncertainty and the integration of features. Inspired by recent advances in deep generative learning, we propose a novel approach leveraging conditional deep generative models (cDGMs) to address these challenges. cDGMs learn the demand distribution and generate probabilistic demand forecasts conditioned on price and features. This generative approach enables accurate profit estimation and supports the design of algorithms for two key objectives: (1) optimizing inventory for arbitrary prices, and (2) jointly determining optimal pricing and inventory levels. We provide theoretical guarantees for our approach, including the consistency of profit estimation and convergence of our decisions to the optimal solution. Extensive simulations—ranging from simple to complex scenarios, including one involving textual features—and a real-world case study demonstrate the effectiveness of our approach. Our method opens a new paradigm in management science and operations research, is adaptable to extensions of the newsvendor and pricing problems, and holds potential for solving other conditional stochastic optimization problems.
}



\maketitle
 
%


\section{Introduction} \label{sec:1}
The classic newsvendor problem has long served as a foundational model in inventory management, offering an effective approach to balancing the costs of understocking and overstocking under demand uncertainty. In practical applications, where demand distributions are often unknown, data-driven approaches \citep{shapiro2003monte} become essential for making informed decisions based on historical data. In addition, demand is often influenced by contextual features—such as customer demographics, seasonality, and other factors. Building on these insights, \citet{ban2019big} formalized the feature-based newsvendor problem, which integrates these features to optimize inventory decisions.

In this paper, we extend the feature-based newsvendor problem to incorporate the influence of price alongside contextual features. Unlike fixed-price scenarios, real-world market prices fluctuate, resulting in historical demand data observed across a range of different prices. This variability introduces two broader scenarios:
\begin{itemize}
\item \textbf{Feature-based Newsvendor with Arbitrary Prices}: Given an arbitrary $p \in \mathcal{P}\in\mathbb{R}^{+}$ and observed features $\bx \in \mathcal{X}$, our goal is to determine the optimal inventory level that maximizes the profit. This corresponds to problem (\ref{eq:oo1}) in Section \ref{sec:3.2}.
\item \textbf{Feature-based Newsvendor Pricing}: In this scenario, given observed features $\bx$, both price and inventory levels are determined jointly to maximize profit. This corresponds to problem (\ref{eq:oo2}) in Section \ref{sec:3.3}. 
\end{itemize}
These two problems are particularly relevant in real-world settings. For instance, in a retail chain, individual store managers set inventory levels based on contextual factors like holidays, weather, and local events, while prices are determined centrally by corporate headquarters, fluctuating due to promotions or market dynamics. In this context, store managers aim to optimize inventory for varying prices and observed features. The second scenario is even more crucial across industries where both price and inventory levels must adapt to features. For example, in e-commerce, aligning pricing and inventory decisions with contextual factors, such as competitor actions and market trends, can significantly enhance profitability.

Several challenges arise in addressing these problems. Existing methods for the feature-based newsvendor typically assume fixed prices, neglecting the impact of price fluctuations on demand. When prices vary, these models require frequent adjustments, which may involve continual retraining or access to comprehensive historical data. Furthermore, the feature-based newsvendor pricing problem introduces significant complexity due to the unknown demand distribution coupled with decision-dependent uncertainty. This problem inherently constitutes a conditional stochastic optimization task with an unknown conditional distribution, rendering commonly used methods, such as empirical risk minimization (ERM), infeasible. Additionally, the integration of contextual features adds further complexity, as the optimal decision becomes a function of these features rather than a fixed point.


In recent years, the explosion of data and advancements in deep learning have transformed numerous fields, with deep neural networks (DNNs) emerging as powerful tools for modeling complex patterns. However, applying DNNs to the feature-based newsvendor pricing problem presents unique challenges. While inventory-only decisions can often be reframed as quantile regression tasks suitable for standard DNNs \citep{oroojlooyjadid2020applying, han2023deep}, the joint pricing and inventory problem does not lend itself to such transformations. This problem remains an optimization over an unknown demand distribution, which renders traditional deterministic deep learning methods inadequate. To overcome this limitation, we recognize the need for a probabilistic approach that can accurately model demand uncertainty, motivating us to shift toward probabilistic deep learning methods.

To address the challenges of demand uncertainty in pricing and inventory optimization, we propose using conditional deep generative models (cDGMs) for probabilistic demand learning. Recent advances in deep generative models (DGMs) have proven effective in capturing complex data distributions by generating samples that reflect the true variability of the data. For example, DGMs are widely used in artificial intelligence applications to generate structured data such as images \citep[DDPM;][]{ho2020denoising} and text \citep[ChatGPT;][]{ouyang2022training}. cDGMs extend DGMs by incorporating contextual data, enabling them to generate samples conditioned on specific inputs. For instance, Stable Diffusion \citep{rombach2022high} can create images from textual descriptions (e.g., “a sunny beach with palm trees”). While cDGMs have demonstrated success in modeling complex data distributions, they are also well-suited to capturing uncertainty in real-valued data structures, such as the demand patterns relevant to our study.

In our approach, the conditional inputs are features and price, with demand as a one-dimensional, real-valued output. Once trained, our cDGMs generate demand samples conditioned on any specified price and features, enabling the estimation of the unknown conditional stochastic optimization objective and facilitating the development of algorithms for optimization problems \eqref{eq:oo1} and \eqref{eq:oo2}. This generative approach directly produces demand samples from the model’s internal structure, eliminating the need to estimate an explicit conditional distribution function or perform numerical integration. This sampling procedure simplifies the optimization process, making it especially efficient for obtaining decisions in both inventory-only and combined pricing-inventory scenarios. In implementing our algorithms, we employ latent variable-based cDGMs, such as variational autoencoders (VAEs) \citep{kingma2013auto} and generative adversarial networks (GANs) \citep{goodfellow2020generative}, while remaining flexible regarding the model type.

We leverage a theoretical result from \citet{zhou2023deep}, which shows that the distribution of samples generated by a cDGM converges to the true conditional distribution of the target. Using this result as a condition, we establish three key theorems that validate our algorithm’s performance. The first theorem shows that, for any given price, our inventory decision is consistent with the true optimal inventory level, confirming the optimality of our approach for the feature-based newsvendor problem across arbitrary prices. This consistency also implies that the newsvendor excess risk, as examined in \citet{han2023deep}, tends to zero. The second theorem focuses on profit estimation: using demand samples generated by our cDGMs, we show that the estimated profit function is consistent with the true conditional expected profit function for any given price. Lastly, the third theorem, along with a corollary, establishes that our joint pricing and inventory decisions are asymptotically optimal, with the resulting profit approaching the true optimal profit. These theoretical results provide rigorous support for the reliability of our cDGM-based approach.

We further validate our proposed method through extensive studies. In a simulation study, we demonstrate the superiority of our method over traditional approaches across a range of data-generating processes, from simple to complex. Additionally, we conduct an experiment incorporating textual features relevant to demand distribution, showing that incorporating such information can improve profitability. Finally, we apply our approach to a real-world dataset on food demand from Kaggle, empirically demonstrating its effectiveness in practical settings.

\subsection{Literature Review}
In this part, we review some related literature.

\subsubsection{Feature-based newsvendor}
Before the formalization of the feature-based newsvendor problem, researchers had already incorporated feature information into newsvendor and inventory control models in various ways. \citet{liyanage2005practical} developed operational statistics to integrate demand estimation and optimization in inventory control. \citet{see2010robust} proposed a robust, factor-based approach for multiperiod inventory management that accounts for trend and seasonality. \citet{beutel2012safety} used causal demand forecasting to account for exogenous factors like price and weather in safety stock planning. \citet{sachs2014data} addressed censored demand observations by incorporating exogenous variables to improve demand accuracy in inventory decisions.

\citet{ban2019big} formally introduced the feature-based newsvendor model, employing empirical risk minimization (ERM) and kernel optimization (KO) to achieve notable cost efficiency improvements. Extending this approach, \citet{oroojlooyjadid2020applying} and \citet{han2023deep} applied deep neural networks (DNNs) to estimate demand quantiles, with \citet{han2023deep} providing theoretical convergence guarantees. More recently, researchers have extensively explored feature-based newsvendor models, applying diverse methods and frameworks to tackle various challenges in the field. For example, \citet{zhang2024optimal} proposed a distributionally robust model with a Lipschitz-regularized framework, enhancing interpretability and empirical performance. \citet{zhao2024private} introduced a privacy-preserving policy using f-differential privacy to secure individual data in feature-based newsvendor models, while \citet{fu2024distributionally} developed a robust solution with a JW ambiguity set to improve resilience in feature-driven demand scenarios. \citet{ding2024feature} proposed adaptive algorithms for censored demand with a dynamic shrinkage approach, achieving enhanced performance in inventory control under demand uncertainty. In a multi-period setting, \citet{qi2023practical} developed an end-to-end feature-based model that reduced inventory costs in e-commerce applications.

In most of these studies, prices remain fixed (or equivalently, the stock-out cost is fixed). The problem \eqref{eq:oo1} considered in our work differs from these settings by allowing variable prices in historical observations, which directly influence the demand distribution. Additionally, our proposed method possesses the capacity to address the inventory decision problem across all prices $p \in \mathcal{P}$ simultaneously. This approach enhances efficiency and offers a convenient foundation for tackling the joint decision problem \eqref{eq:oo2}.

\subsubsection{Joint Inventory and Pricing Decisions}
The newsvendor problem with pricing decisions has been extensively studied since \citet{whitin1955inventory}. Early studies often assume complete information about demand and employ specific demand structures, such as additive \citep{mills1959uncertainty}, multiplicative \citep{kincaid1963inventory}, or mixed forms \citep{young1978price}. For a comprehensive overview of these models, see \citet{qin2011newsvendor} and \citet{deyong2020price}. However, these models are often limited in practice, as real-world demand is typically more complex than assumed and its precise form is unknown.

The data-driven settings are increasingly relevant in today’s rapidly evolving environment and attracting significant research interest. In such scenarios without specific demand structures, the joint decision problem becomes a conditional stochastic optimization, requiring specialized algorithms. In such cases, the prescriptive approach by \citet{bertsimas2020predictive} offers a feasible by using machine learning techniques to estimate the objective through weighted historical samples. \citet{harsha2021prescriptive} designs tailored linear models to estimate the objective in a data-driven price-setting newsvendor. \citet{liu2023solving} proposes an approximate gradient descent (AGD) algorithm that employs weighted historical samples to approximate the gradients, aiming to solve the optimization issue encountered in \citet{bertsimas2020predictive}. Our method for \eqref{eq:oo2} differs by leveraging deep neural networks (DNNs), which excel in handling high-dimensional, complex data, enhancing performance in data-rich environments. This approach is efficient, requiring only a single training process and enabling decision-making without retraining or ongoing access to historical data—unlike the weighted estimations in \citet{bertsimas2020predictive}, which require continuous data access. This advantage is essential for large-scale applications and benefits data privacy by minimizing repeated access to sensitive information.

Another line of research explores joint inventory and pricing in multi-period or dynamic settings, typically without incorporating observed features. For example, \citet{chen2019coordinating} proposed a data-driven approach to dynamically manage pricing and inventory over a finite planning horizon, learning demand patterns to set optimal prices with minimal profit loss. \citet{qin2022data} developed an approximation algorithm for multi-period settings that leverages historical demand data to support near-optimal decision-making. More recently, \citet{zheng2024dual} applied a dual-agent reinforcement learning model to manage dynamic pricing and replenishment with varying decision frequencies for inventory and pricing. Although our study focuses on single-period newsvendor pricing, it is relevant to dynamic, multi-period problems and has potential for future extensions. For example, our joint decision-making algorithm for \eqref{eq:oo2}, adaptable to any observed feature $\bx$, could be effectively deployed over extended periods, even though multi-period dynamics are not the primary focus here and could be explored further in future work.

\subsubsection{Deep Generative Models} 
There are several categories of deep generative models (DGMs), such as autoregressive and latent variable models, with a detailed taxonomy provided by \citet{Tomczak202DeepGenerative}. Popular large language models (LLMs), such as GPT-4 \citep{achiam2023gpt}, predominantly use an autoregressive approach, capitalizing on semantic context within sentences. For tasks like image generation, state-of-the-art methods often use structures like diffusion models \citep[e.g.,][]{ho2020denoising}, which base on iterative denoising processes that gradually transform noise into clear images. Classic generative approaches include GANs, VAEs, and flow-based models \citep[e.g.,][]{dinh2014nice}, each offering distinct frameworks for data generation with unique mechanisms. These models and their combinations have also been widely studied to leverage their complementary strengths.

Recently, cDGMs have demonstrated significant potential in probabilistic forecasting for real-valued data. Starting from \citet{rasul2021autoregressive}, cDGMs have gained attention as powerful tools for probabilistic time series forecasting \citep[e.g.,][]{li2022generative}. In parallel, \citet{han2022card} extended cDGMs to support regression and classification tasks, improving uncertainty estimation over traditional methods. In theoretical advances for conditional generation, \citet{liu2021wasserstein} established non-asymptotic error bounds and \citet{zhou2023deep} demonstrated consistency for conditional sampling in high-dimensional contexts. Additionally, theoretical work on unconditional DGMs may offer insights applicable to cDGMs \citep[e.g.,][]{dahal2022deep, oko2023diffusion, chae2023likelihood}.

\subsection{Organization}
The remainder of this paper is organized as follows. Section \ref{sec:2} provides preliminaries on the newsvendor problem. Section \ref{sec:3} introduces the feature-based newsvendor and pricing problems, formulates the two objectives \eqref{eq:oo1} and \eqref{eq:oo2} studied in this paper. In Section \ref{sec:4}, we introduce cDGMs and the proposed algorithms for solving the concerning objectives, along with some analysis of related methods. Section \ref{sec:5} presents three theoretical results supporting the effectiveness of our proposed methods. Experiments using simulated and real-world datasets are detailed in Sections \ref{sec:6} and \ref{sec:7}, respectively. Finally, Section \ref{sec:8} concludes the paper and discusses potential directions for future research. Proofs and additional materials are provided in the Appendices.

\section{Preliminaries: Newsvendor Problem} \label{sec:2}
In this section, we briefly introduce the single-period newsvendor problem along with its feature-based extension. Consider a retailer selling a product over a single sales period, where demand \( D \) is uncertain and modeled as a random variable with distribution \( F_{D}(\cdot) \). The retailer needs to decide on an order quantity \( q \in \mathbb{R}^+ \), representing the units of the product to be ordered. Let $p$ denote the unit retail price, $c$ the unit purchase cost, and $s$ the unit salvage value when there is excess inventory. The profit function, given demand $d$, price $p$, and order quantity $q$, is defined as 
\begin{align} 
    \Pi(d, p, q) &= p \min (q, d) + s \max (q - d, 0) - c q \notag\\ 
    &= (p - c) d - (c - s)(q - d)^+ - (p - c)(d - q)^+,\notag
\end{align}
where $(x)^+=\max(x,0)$ denotes the positive part of $x$.

\subsection{Basic Newsvendor} \label{sec:2.1}
In a basic newsvendor problem, the price \( p \) is fixed, and the retailer must decide on $q$ before demand is realized, with the goal of maximizing expected profit, given by
\begin{align} 
\pi_D(q) &:= \mathbb{E}[\Pi(D, p, q)],\notag
\end{align}
where the expectation is taken with respect to the random demand \( D \). Defining \( c_o = c - s \) (unit overage cost) and \( c_u = p - c \) (unit underage cost), we can frame this as an inventory problem where $q$ represents the inventory level. The optimal \( q \) corresponds to a quantile of the demand distribution \( F_D \):
\begin{align} \label{eq:problem}
q^* &= \underset{q \geq 0}{\operatorname{argmax}} \ \pi_D(q) = \underset{q \geq 0}{\operatorname{argmin}}\ c_o \mathbb{E}[(q - D)^{+}] + c_u \mathbb{E}[(D - q)^{+}] \notag\\
&= \inf \{q \geq 0: F_{D}(q) \geq \rho\}, \quad \text{where } \rho := \frac{c_u}{c_u + c_o}.
\end{align}
Thus, in data-driven settings, the optimal inventory decision $q*$ is achieved by estimating the \( \rho \)-quantile of demand \( D \).

\subsection{Feature-Based Newsvendor} \label{sec:2.2}
The feature-based newsvendor model extends the decision on \( q \) by incorporating contextual features \( \mathbf{X} \in \mathbb{R}^{k} \) \citep{ban2019big}. Since these features can influence demand distribution, inventory decisions should adapt based on the observed features \( \mathbf{X} = \mathbf{x} \). Therefore, the optimal decision \( q \) is expressed as a function \( q = q(\mathbf{x}) \), modifying problem (\ref{eq:problem}) to optimize:
\begin{align} \label{eq:q*}
    q_\rho(\mathbf{x}) & := \underset{q\in\mathbb{R}^+}{\arg\max} \ \mathbb{E}_{D | \mathbf{X}}[\Pi(D, p, q(\mathbf{X})) | \mathbf{X} = \mathbf{x}], \notag\\
    & = \underset{q\in\mathbb{R}^+}{\arg\min} \ \mathbb{E}_{D | \mathbf{X}}[ c_o(q(\mathbf{X})-D)^+ + c_u(D - q(\mathbf{X}))^+  | \mathbf{X} = \mathbf{x}].
\end{align}
The optimal solution for (\ref{eq:q*}) is given by the conditional quantile function
\begin{align} \label{eq:frho}
    q_\rho(\mathbf{x}) = \inf \{q \geq 0 : F_{D|\mathbf{X}}(q | \mathbf{x}) \geq \rho\}.
\end{align}
In data-driven scenarios, where the demand distribution is unknown, \( q_\rho \) must be estimated from historical observations, denoted by \( \tilde{S}_n = \{(\mathbf{x}_i, d_i)\}_{i=1}^{n} \). The task is to learn an approximate function \( \hat{q}(\mathbf{x}) \) that aims to match \( q_{\rho}(\mathbf{x}) \).

\section{Feature-based Newsvendor and Pricing Problem} \label{sec:3}
In practical applications, demand is often price-sensitive and may vary within a range rather than remaining fixed, as assumed in Sections \ref{sec:2}. This introduces a setting where both demand and the optimal inventory decision depend on price. We extend the newsvendor model in Section \ref{sec:2} by allowing price to take values from a set \( \mathcal{P}\subset\mathbb{R}^+ \), which may be either continuous or discrete. We assume that $\mathcal{P}$ is a bounded set with an upper bound $p_{\max}$.

In this framework, for each sales period, the retailer observes features $\mathbf{X}$ with support $\mathcal{X}$  which can contain both continuous and categorical components. Our goal is to maximize profit based on these observed features. This section introduces the demand modeling approach used in this study, followed by two decision-making scenarios we mainly focus on.

\subsection{Demand Modeling with Features and Price} \label{sec:3.1}
For demand modeling, traditional studies often assume a demand function \( D(p, \epsilon) \) \citep{huang2013demand}, where demand depends on price \( p \) and a price-independent random variable \( \epsilon \). In contrast, we model demand as a conditional distribution dependent on both price and features. When treating price as a random variable, we denote it as $P$. This approach avoids assuming a specific functional form between \( D \) and \( P \), making this framework adaptable to a wide range of stochastic demand models in demand learning (see Table 2 in \cite{huang2013demand} for a comprehensive summary). For example, a model where \( D = \mu(p) + \epsilon \), with deterministic \( \mu(p) \) and random \( \epsilon \), is a special case where \( D | (P = p) \) has a cumulative distribution function (CDF) \( F(d|p) = F_{\epsilon}(d - \mu(p)) \), where \( F_{\epsilon}(\cdot) \) denotes the cumulative distribution function of \( \epsilon \).

To further generalize, we incorporate features as described in Section \ref{sec:2.2}. Specifically, we assume the conditional density function of \( D|(\mathbf{X} = \mathbf{x}, P = p) \) is \( f_{D|\mathbf{X},P}(\cdot | \mathbf{x}, p) \). We define the conditional expected profit function as
\begin{align} \label{eq:pipqx}
    \pi(\mathbf{x}, p, q) &= \int_{-\infty}^{\infty} \Pi(t, p, q) f_{D|\mathbf{X},P}(t | \mathbf{x}, p) \, dt \notag \\
    &= \mathbb{E}_{D | \mathbf{X}, P} \big\{ \Pi\big(D, p, q \big) \big| \mathbf{X} = \mathbf{x}, P = p \big\},
\end{align}
where the expectation is taken with respect to \( D | (\mathbf{X} = \mathbf{x}, P = p) \). This conditional expected profit function $\pi(\bx,p,q)$ is central to our study, serving as the objective for the decision problems defined and analyzed in the following sections and as the primary function we aim to estimate with our proposed methods.

\subsection{Feature-Based Newsvendor with Arbitrary Prices} \label{sec:3.2}
Given the conditional profit function $\pi(\mathbf{x}, p, q)$ established in Section \ref{sec:3.1}, our objective is to maximize expected profit by determining an optimal inventory level $q$ for each observed feature-price pair $(\mathbf{x}, p)$. Unlike the traditional newsvendor model, this requires solving a decision problem where price varies, introducing unique quantile levels for each $p$.

To optimize profit, we define the optimal decision function pointwise for any \( (\mathbf{x}, p) \in \mathcal{X} \times \mathcal{P} \) as
\begin{align} \label{eq:oo1}
    q^*(\mathbf{x}, p) := \underset{q \in \mathbb{R}^+}{\arg\max} \, \pi(\mathbf{x}, p, q),  \tag{O-A}
\end{align}
which, through simple derivations, takes the form similar to (\ref{eq:frho}):
\begin{align} 
    q^*(\mathbf{x}, p) = \inf \Big\{ q \geq 0 : F_{D|\mathbf{X},P}(q | \mathbf{x}, p) \geq \frac{p - c}{p - s} \Big\}, \notag
\end{align}
where \( F_{D|\mathbf{X},P}(\cdot | \mathbf{x}, p) \) denotes the conditional CDF.

In a data-driven scenario, we lack information about the underlying data distribution $F_{D|\mathbf{X},P}$. Instead, we rely on historical records from \( n \) sales periods, represented as \( S_n = \{(\mathbf{x}_i, p_i, d_i)\}_{i=1}^n \), where \( \mathbf{x}_i \in \mathcal{X} \) denotes the feature vector, \( p_i \in \mathcal{P} \) the price, and \( d_i \) the observed demand. The problem (\ref{eq:oo1}) can thus be treated as a feature-based newsvendor problem with variable pricing.

Existing methods, particularly those focused on pure inventory optimization, offer valuable insights for inventory decisions at a fixed price. However, in scenarios where price $p\in\mathcal{P}$ varies, each price introduces a distinct quantile level $(p-c)/(p-s)$, making these methods less convenient for repeated application across multiple prices. Below, we analyze the most widely used methods for solving \eqref{eq:oo1}.

\paragraph{Sample Average Approximation (SAA).} SAA estimates $q^*(\bx,p)$ by applying the quantile level $\rho(p)=(p-c)/(p-s)$ for each price $p$. However, this approach has significant drawbacks: it lacks flexibility for prices not observed in the dataset $S_n$ and does not account for feature information, leading to poor performance in our problem.

\paragraph{Empirical Risk Minimization (ERM).} ERM methods estimate the conditional $\rho(p)$-th quantile of $D$ given $\bX,P$, with $\rho(p)$ adjusted per price. For each price $p$ ERM requires a seperate model:
\begin{align}
    \hat{q}_{\operatorname{ERM},p}(\cdot,\cdot) = \underset{q\in \mathcal{F}}{\arg\min} \sum_{i=1}^n \left[(c-s)(q(\bx_i,p_i)-d_i)^+ + (p-c)(d_i - q(\bx_i,p_i))\right], \notag
\end{align}
where $\mathcal{F}:=\{f:\mathcal{X}\times\mathcal{P}\rightarrow\mathbb{R}^+\}$ represents a certain class of regression models. The decision function is then given by $\hat{q}_{\operatorname{ERM}}(\bx,p) := \hat{q}_{\operatorname{ERM},p}(\bx,p)$. When $\mathcal{F}$ represents the class of linear models, we refer to this as ERM-LR \citep{ban2019big}, and when $\mathcal{F}$ represents the class of neural networks, as ERM-NN \citep{han2023deep}. 

\paragraph{Kernel-Weights Optimization (KO).} With a slight modification to the original KO method in \cite{ban2019big}, the Nadaraya–Watson estimator can be used to account for both features and price using kernel functions \( K_H(\cdot) \) and \( K_h(\cdot) \) to estimate $\pi(\bx,p,q)$:
\begin{align} 
    \hat{\pi}_{\operatorname{KO}}(\bx,p,q) = \frac{\sum_{i=1}^nK_{H}(\bx-\bx_i)K_{h}(p-p_i)\Pi(d_i,p_i,q)}{\sum_{i=1}^nK_{H}(\bx-\bx_i)K_{h}(p-p_i)}. \notag
\end{align}
The decision is then given by
\begin{align}
    \hat{q}_{\operatorname{KO}}(\bx,p) = \underset{q\in \mathbb{R}^+}{\operatorname{argmax}}\ \hat{\pi}_{\operatorname{KO}}(\bx,p,q). \notag
\end{align}

\paragraph{Residual-based Estimation (RBE).} 
Assuming an additive demand structure $D|(\bX=\bx,P=p)=\alpha p+\beta^T\bx+\epsilon$, RBE seperates demand into a linear component and noise. We estimate $\alpha$ and $\beta$, then calculate residuals $\hat{\mathcal{E}}=\{\hat{\epsilon}_i:=d_i-\hat{\alpha} p_i-\hat{\beta}^T\bx_i\}_{i=1}^n$. The estimated of $q^*(\bx,p)$ is given by
\begin{align}
    \hat{q}_{\operatorname{RBE}}(\bx,p) =\hat{\alpha}p + \hat{\beta}^T\bx + \inf\left\{\epsilon:\hat{F}_{\hat{\mathcal{E}}}(\epsilon)\geq \frac{p-c}{p-s} \right\}, \notag
\end{align}
where $\hat{F}_{\hat{\mathcal{E}}}(\epsilon) = \frac{1}{n} \sum_{i=1}^{n} \mathbb{I}(\hat{\epsilon}_i \leq \epsilon)$ is the empirical CDF based on $\hat{\mathcal{E}}$. 
This method resembles decision-maker 2 (DM2) approach approach in Theorem 3 of \citet{ban2019big}. While simple and fast, it can underperform if the additive demand model does not accurately represent the demand structure.

\subsection{Feature-Based Newsvendor Pricing} \label{sec:3.3}
We now present a general framework for joint pricing and inventory decisions under the feature-based newsvendor model. In this scenario, the retailer decides both the price $p$ and order quantity $q$ simultaneously at the start of the sales period, based on the observed feature vector $\bx$. We define the optimal decision functions \( p_{\operatorname{opt}}(\cdot) \) and \( q_{\operatorname{opt}}(\cdot) \) pointwise for any \( \mathbf{x} \in \mathcal{X} \) as
\begin{align} \label{eq:oo2}
    (p_{\operatorname{opt}}(\mathbf{x}), q_{\operatorname{opt}}(\mathbf{x})) 
    & := \underset{(p, q) \in \mathcal{P} \times \mathbb{R}^+}{\operatorname{argmax}} \ \pi(\mathbf{x}, p, q). \tag{O-B} 
\end{align}

\begin{figure}
    \centering
    \includegraphics[width=0.5\linewidth]{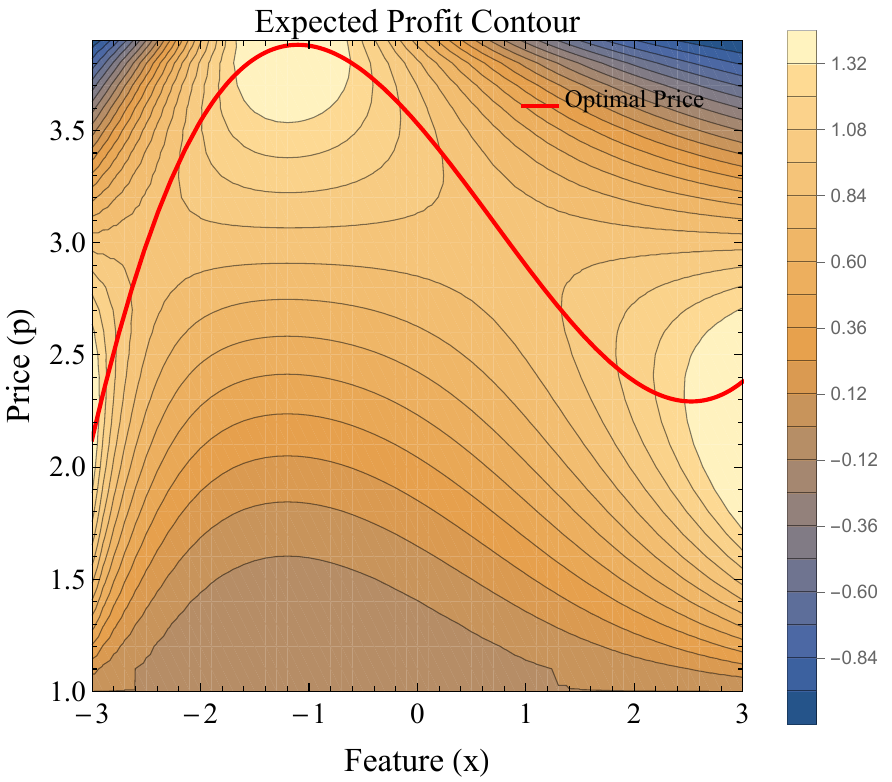}
    \caption{The contour plot of $\pi(x,p,q^*(\bx,p))$ for model (a) in Example \ref{example:2} across Feature $\bx$ and Price $p$ with a red line indicating the optimal pricing path for maximum profit.}
    \label{fig:contour}
\end{figure}

To illustrate how incorporating features complicates the traditional price-setting newsvendor problem, consider the following example.
\begin{example} \label{example:2}
    Let $k=1$ and consider two demand models: 
    \begin{align} 
        \text{(a)}\ D|(\bX=\bx,P=p)= (4-p)^{1/6} + \epsilon; \quad \text{(b)}\ D|(\bX=\bx,P=p)= (4-p)^{h(\mathbf{x})} + \epsilon, \notag
    \end{align}
    where $h(\bx)=(6+5\bx)^2/(216+48\bx)+0.01$, $\epsilon\sim \mathcal{N}(0,1)$. 
\end{example}
In model (a), demand does not depend on features, which is a special case of model (b) with $\bx=0$. If we have access to the demand distribution and the optimal inventory function $q^*(\bx,p)$, we can visualize the expected profit $\pi(\bx,p,q^*(\bx,p))$ for model (b) by a contour plot. As shown in Figure \ref{fig:contour}, the red line  represents the optimal price for varying values of $\bx$, revealing significant variation across feature values. 
This example demonstrates that, without features, the optimal pricing is simply a fixed scalar value. In contrast, feature-based optimal pricing requires finding a function of
$\bx$, making two tasks differ greatly in complexity.

In the data-driven scenario, we rely on historical data \( S_n = \{(\mathbf{x}_i, p_i, d_i)\}_{i=1}^n \) to derive the optimal decisions. Unlike the inventory-only decision problem, where an explicit quantile-based solution is available as in Section \ref{sec:3.2}, no closed-form solution exists for the joint decision problem (\ref{eq:oo2}), even for simplified demand models such as additive or multiplicative demand \citep{petruzzi1999pricing}. This prevents us from converting the task into a straightforward prediction problem, as we could with inventory-only decisions in (\ref{eq:oo1}). Consequently, one of the primary challenges lies in estimating the unknown objective function $\pi(\bx,p,q)$ to address the joint decision-making problem.

Several challenges arise looking for feasible solutions from standard methods in the newsvendor problem to solve (\ref{eq:oo2}). The objective function’s uncertainty is decision-dependent because price influences the demand distribution. This dependency causes ERM methods to give incorrect estimations of  \(\pi(\bx, p, q)\). To illustrate, if we treat the empirical negative profit as empirical risk, the resulting optimization problem over decision functions $(q(\cdot),p(\cdot))$ is
\begin{align} 
    \underset{q,p\in\mathcal{F}}{\min} \hat{R}(q(\cdot),p(\cdot)) = -\frac{1}{n}\sum_{i=1}^n \Pi(d_i,p(\bx_i),q(\bx_i)).  \notag
\end{align}
The empirical risk $\hat{R}$ yields a value of negative infinity when $p(\bx)\equiv\infty$, rendering it practically meaningless. The issue arises because changes in price lead to corresponding, yet unobservable, changes in demand.

Secondly, using a deterministic demand predictor does not yield accurate estimates for $\pi(\bx,p,q)$. To see this, consider an oracle demand predictor defined as \( d_{\mu}(\mathbf{x}, p) = \mathbb{E}_{D | \mathbf{X}, P}(D | \mathbf{X} = \mathbf{x}, P = p) \), which is the optimal point predictor of \( D \) in terms of minimizing expected squared error. Estimating \( \pi(\mathbf{x}, p, q) \)  using \( d_{\mu}(\mathbf{x}, p) \) yields
\begin{align} \label{eq:pimu}
    \pi_{\mu}(\mathbf{x}, p, q) = \Pi(d_{\mu}(\mathbf{x}, p), \mathbf{x}, p),
\end{align}
which leads to the following result:
\begin{proposition} \label{prop:2}
    For any \( (\mathbf{x}, p, q) \in \mathcal{X} \times \mathcal{P} \times \mathbb{R}^+ \),
    \begin{align}
        \pi_\mu(\mathbf{x}, p, q) \geq \pi(\mathbf{x}, p, q), \notag
    \end{align}
    with equality only if \( D | (\mathbf{X} = \mathbf{x}, P = p) \) is a degenerate distribution. 
\end{proposition}
Proposition \ref{prop:2} shows that \( \pi_{\mu}(\mathbf{x}, p, q) \) generally overestimates \( \pi(\mathbf{x}, p, q) \), except in trivial cases, indicating an inaccurate estimation even when the true mean demand is known. Note that  $\arg\max_{q\in\mathbb{R}^+} \pi_\mu(\mathbf{x}, p, q) = d_{\mu}(\mathbf{x}, p)$, which does not match the optimal solution $q^*(\bx,p)$ for problem \eqref{eq:oo1}. This implies that it can not provide correct decision for problem \eqref{eq:oo2}. Consequently, regardless of how accurately models predict $d_{\mu}(\mathbf{x}, p)$, optimizing based on \eqref{eq:pimu} is fundamentally misspecified, leading to inconsistent decisions. Deterministic prediction models, therefore, prove inadequate as they fail to capture the uncertainty in the objective function. This limitation motivates a shift toward probabilistic prediction models, specifically generative models, which we introduce in Section \ref{sec:4}.

Finally, an existing feasible approach within the standard paradigm is the prescriptive method proposed by \cite{bertsimas2020predictive}. This method assigns data-driven weights $w_i(\bx,p)$ to the historical observations and optimizes the objective by solving
\begin{align} \label{eq:presc}
    \hat{q}_{\operatorname{ML}}(\bx),\hat{p}_{\operatorname{ML}}(\bx)=\underset{q,p\in \mathbb{R}^+}{\operatorname{argmax}}\  \sum_{i=1}^N w_i(\bx,p) \Pi(d_i,p,q).
\end{align}

\section{Conditional Deep Generative Models Solutions} \label{sec:4}
In this section, we address the challenges in optimization problems (\ref{eq:oo1}) and (\ref{eq:oo2}) using cDGMs. First, we introduce latent variable cDGMs and explain how these models capture the conditional distribution of $D|\bX,P$. Next, we present estimation methods for the objective function $\pi(\bx,p,q)$ based on demand samples generated by cDGMs and outline decision-making algorithms for solving the optimization problems. Finally, we provide a comparison of methods.

\subsection{Latent Variable Conditional Deep Generative Models}
We introduce latent variable DGMs and their conditional extensions, clarifying the application of cDGMs to our specific problem. To closely align with our study objectives, we use context-specific notations rather than conventional symbols from deep learning literature.

In latent variable DGMs, noise $\boldsymbol{\eta} \in \mathbb{R}^r$ plays a central role as a latent variable, serving as the foundation for generating new data. Models such as GANs, VAEs, and denoising diffusion probabilistic models \citep[DDPMs;][]{ho2020denoising}, use a random noise vector sampled from a standard Gaussian distribution.
Our focus is on modeling the distribution of demand \( D \in \mathbb{R}^+ \). Rather than modeling the distribution function of $D$ explicitly, generate artificial samples that approximate the distribution of $D$. For unconditional generation, a function map $g:\mathbb{R}^r\rightarrow \mathbb{R}^+$ is required to map any random noise $\boldsymbol{\eta} \sim \mathcal{N}(\mathbf{0}, \mathbf{I}_r)$ to the target via $d=g(\boldsymbol{\eta})$. In our problem, we require the conditional distribution $D|\bX,P$ rather than the unconditional one, necessitates a function map with condition $(\bx,p)$ as input,  $G: \mathcal{X} \times \mathcal{P} \times \mathbb{R} \rightarrow \mathbb{R}^+$ with the aim that $G(\bx,p,\boldsymbol{\eta})$ follows the similar distribution of $D|(\bX=\bx,P=p)$.

Before detailing the modeling of such a function map, a key question arises: does a function $G$ exist such that this approximation is sufficiently accurate? This question is addressed by \citet{zhou2023deep} through a noise-outsourcing lemma.
\begin{lemma} \label{lemma:1}
Let $(\bX, P, D)$ be a random triplet taking values in $\mathcal{X}\times\mathcal{P} \times \mathbb{R}$ with joint distribution $f_{\mathbf{X},P, D}$. Then there exist a random vector $\boldsymbol{\eta} \sim \mathcal{N}\left(\mathbf{0}, \mathbf{I}_r\right)$ for any given $r \geq 1$ and a Borel-measurable function $G:  \mathcal{X}\times\mathcal{P} \times \mathbb{R}^r \rightarrow \mathbb{R}$ such that $\boldsymbol{\eta}$ is independent of $(\bX,P)$ and
\begin{align} \label{eq:zd}
    (\bX,P, D)=(\bX,P, G(\bX,P, \boldsymbol{\eta})) \text { almost surely. }
\end{align}
\end{lemma}
A direct consequence of Lemma \ref{lemma:1} is that if \( G^* \) satisfies (\ref{eq:zd}), then \( G^*(\mathbf{x}, p, \boldsymbol{\eta}) \) follows the conditional distribution \( D | (\mathbf{X} = \mathbf{x}, P = p) \), verifying the existence of such a \( G^* \). 

Deep learning methods can be implemented to model and learn the function map. Specifically, we apply a deep neural network \( G_{\theta}(\cdot, \cdot, \cdot) : \mathcal{X} \times \mathcal{P} \times \mathbb{R} \rightarrow \mathbb{R}^+ \), parameterized by \( \theta \), which we refer to as the generator.

\begin{figure}
    \centering
    \includegraphics[width=1.0\linewidth]{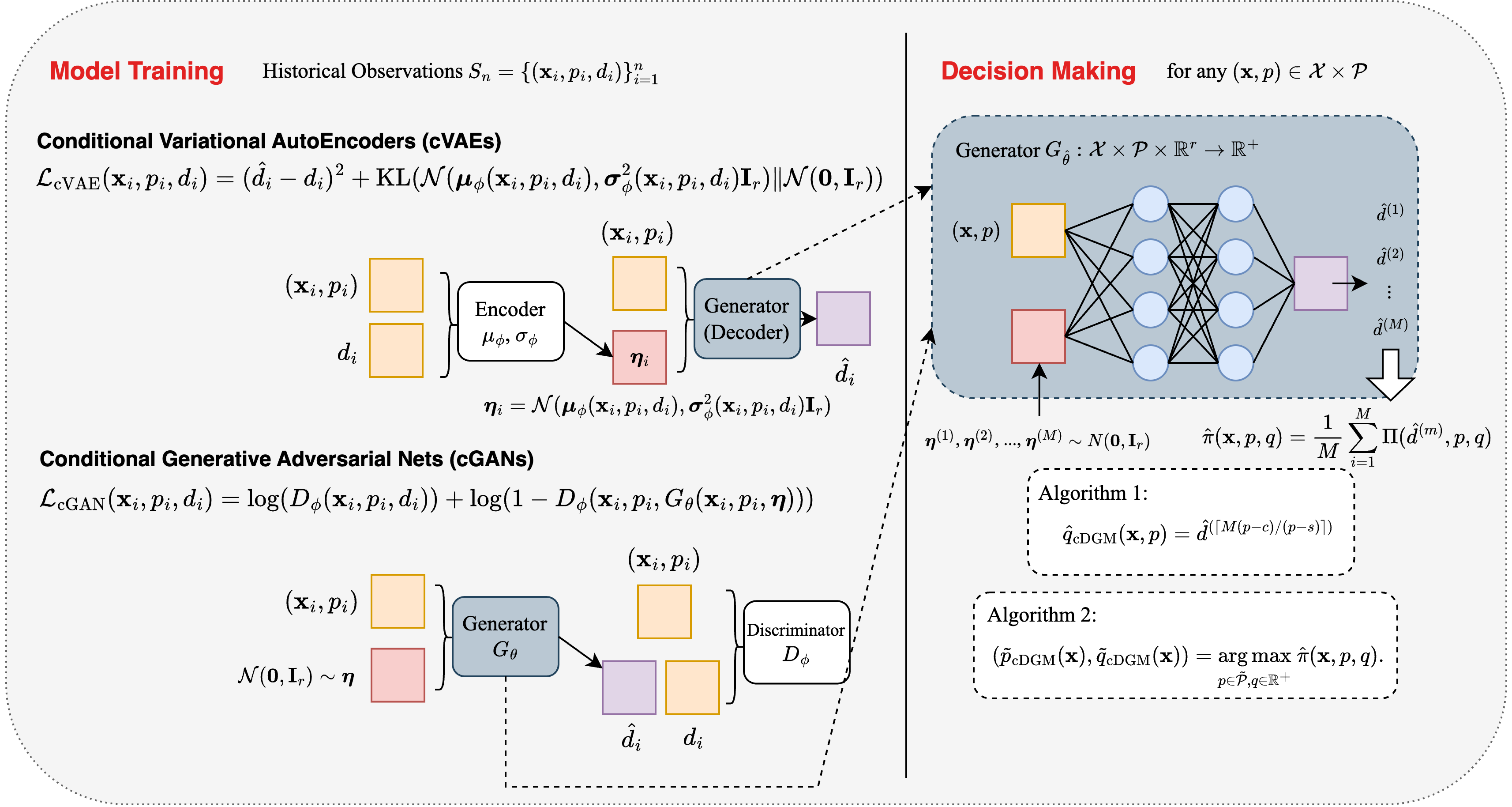}
    \caption{Examples of cDGMs and the Decision Making Procedure}
    \label{fig:generative}
\end{figure}

Based on the historical data $S_n$, we can learn the generator $G_{\hat{\theta}}$. The learning processes for obtaining $G_{\hat{\theta}}$  across different types of cDGMs vary. In practical applications, model selection is unrestricted; the choice can depend on data complexity and cross-validation performance. In Figure \ref{fig:generative}, we illustrate two alternative methods: conditional VAEs (cVAE) and conditional GANs (cGAN) in the left panel. Detailed mathematical formulations are provided in Appendix C.
\subsection{Decision-Making with cDGMs}
For any \( (\bx,p)\in\mathcal{X}\times\mathcal{P} \), the generator \( G_{\hat{\theta}} \) can produce an arbitrary number of artificial demand samples as follows:
\begin{align} \label{eq:generator}
    \boldsymbol{\eta}^{(m)} \sim \mathcal{N}(\mathbf{0}, \mathbf{I}), \quad \hat{d}^{(m)} = G_{\hat{\theta}}(\mathbf{x}, p, \boldsymbol{\eta}^{(m)}), \quad m = 1, 2, 3, \dots.
\end{align}
We denote by \( \hat{S}(\bx,p) = \{\hat{d}^{(m)}\}_{m=1}^M \)  the set of \( M \) generated demand samples according to \eqref{eq:generator}. Using these samples, we construct a direct estimator of the conditional expected profit function \( \pi(\mathbf{x}, p, q) \) as:
\begin{align}  \label{eq:dgmest}
    \hat{\pi}(\bx,p,q) = \frac{1}{M}\sum_{m=1}^M\Pi(\hat{d}^{(m)},p,q).
\end{align}
Ideally, if \( G_{\hat{\theta}} \) closely approximates the true generator in Lemma \ref{lemma:1} and \( M \) is sufficiently large, we obtain:
\begin{align}
     \hat{\pi}(\bx,p,q) \approx \mathbb{E}_{ \{\boldsymbol{\eta}^{(m)}\}_{m=1}^M \sim \mathcal{N}(\mathbf{0},\mathbf{I})} (\hat{\pi}(\bx,p,q)) \approx \pi(\mathbf{x}, p, q), \notag
\end{align}
indicating that \( \hat{\pi}(\bx,p,q) \) can serve as a good approximation of the true objective function.

To solve problem \eqref{eq:oo1}, we optimize \( \hat{\pi}(\mathbf{x}, p, q) \) with respect to $q$:
\begin{align} \label{eq:solveo1}
    \hat{q}_{\operatorname{cDGM}}(\mathbf{x}, p) = \underset{q \in \mathbb{R}^+}{\argmax} \ \hat{\pi}(\mathbf{x}, p, q).
\end{align}
For the joint decision problem \eqref{eq:oo2}, we similarly optimize:
\begin{align} \label{eq:solveo2}
    (\hat{p}_{\operatorname{cDGM}}(\mathbf{x}), \hat{q}_{\operatorname{cDGM}}(\mathbf{x})) = \underset{p \in \mathcal{P}, q \in \mathbb{R}^+}{\argmax} \ \hat{\pi}(\mathbf{x}, p, q).
\end{align}
Since estimating \( \hat{\pi}(\mathbf{x}, p, q) \) requires only forward passes through the neural network generator, this approach is computationally efficient. However, the estimated function \( \hat{\pi}(\mathbf{x}, p, q) \) may be non-convex or exhibit discontinuities, potentially complicating the optimization steps in \eqref{eq:solveo1} and \eqref{eq:solveo2}. We address these challenges and introduce algorithms to streamline the optimization process, as summarized in Figure \ref{fig:generative}(right panel). In what follows, we detail these procedures.

\subsubsection{Algorithm for Optimal Inventory (\ref{eq:oo1})} \label{sec4.2.1}
Due to the simple form of the estimator $\hat{\pi}(\bx,p,q)$, the optimal solution for \eqref{eq:solveo1} can be derived as follows:
\begin{proposition} \label{prop:1}
    Let \( \hat{S}(\bx,p)=\{\hat{d}^{(m)}\}_{m=1}^M \) be sorted in ascending order. The solution to \eqref{eq:solveo1} is given by $ \hat{q}_{\operatorname{cDGM}}(\mathbf{x}, p) =\hat{d}^{\left(\left\lceil M(p - c) / (p - s) \right\rceil\right)} $.
\end{proposition}
Proposition \ref{prop:1} implies that the optimal order quantity \( \hat{q}_{\operatorname{cDGM}}(\mathbf{x}, p) \) corresponds to the \( (p - c) / (p - s) \)-th empirical quantile of the generated demand set \( \hat{S}(\bx,p) \). This result eliminates potential optimization challenges, as the optimal solution is given directly by a quantile of the generated samples. The process of generating samples and obtaining inventory decision is outlined in Algorithm \ref{algo:0}.

\begin{algorithm}[htb]
\caption{Demand Generation and Inventory Decision}\label{algo:0}
\begin{algorithmic}[1] 
    \State \textbf{Input:} Trained cDGM $G_{\hat{\theta}_n}(\cdot,\cdot,\cdot)$, parameters $c$ and $s$, price \( p \) and feature vector \( \bx \) for the sales period, the number of generated samples $M$.
    \State Generate a demand set $\hat{S}(\bx,p)=\{\hat{d}^{(m)}\}_{m=1}^M$ of size $M$ by
    \begin{align}
        \boldsymbol{\eta}^{(m)} \sim \mathcal{N}(0, \mathbf{I}_r),\quad \hat{d}^{(m)} = G_{\hat{\theta}_n}(\bx,p,\boldsymbol{\eta}^{(m)})\quad (m=1,...,M).\notag
    \end{align}
    \State Set quantile level $\rho(p)\leftarrow(p-c)/(p-s)$. 
    \State Obtain inventory decision $\hat{q}(\bx,p)\leftarrow $ $\rho(p)$-th empirical quantile of $\hat{S}(\bx,p)$.
    \State \textbf{Output:} order quantity $\hat{q}(\bx,p)$, generated demands $\hat{S}(\bx,p)$.
\end{algorithmic}
\end{algorithm}

\subsubsection{Algorithm for Optimal Pricing and Inventory (\ref{eq:oo2})} \label{sec:4.2.2}
Without a closed-form solution as in \eqref{eq:solveo1}, optimizing \eqref{eq:solveo2} is challenging. To address this, we simplify \eqref{eq:solveo2} by discretizing the price set. Specifically, we replace the price set $\mathcal{P}$ with a discrete approximation \( \tilde{\mathcal{P}} = \{p_j\}_{j=1}^J \). If \( \mathcal{P} \) is already discrete, we set \( \tilde{\mathcal{P}} = \mathcal{P} \). The resulting optimization problem becomes:
\begin{align} 
    (\tilde{p}_{\operatorname{cDGM}}(\mathbf{x}), \tilde{q}_{\operatorname{cDGM}}(\mathbf{x})) &= \underset{p \in \tilde{\mathcal{P}}, q \in \mathbb{R}^+}{\argmax} \ \hat{\pi}(\mathbf{x}, p, q).  \notag
\end{align}
By leveraging our solution to problem \eqref{eq:solveo1}, we can further simplify this optimization as:
\begin{align} \label{eq:tildep}
    \tilde{p}_{\operatorname{cDGM}}(\mathbf{x}) &= \underset{p \in \mathcal{P}}{\argmax} \ \hat{\pi}(\mathbf{x}, p, \hat{q}_{\operatorname{cDGM}}(\mathbf{x}, p)), \\
    \tilde{q}_{\operatorname{cDGM}}(\mathbf{x}) &= \hat{q}_{\operatorname{cDGM}}(\mathbf{x}, \tilde{p}_{\operatorname{cDGM}}(\mathbf{x})). \notag
\end{align}
This approach is computationally efficient, as we can generate demand samples in parallel for all  \( p \in \tilde{\mathcal{P}} \), taking advantage of matrix computations inherent to neural networks. The entire procedure is outlined in Algorithm \ref{algo:1}. Although we approximate $\mathcal{P}$ by discretization, selecting a sufficiently dense $\tilde{\mathcal{P}}$ enables a close approximation to the optimal solution of \eqref{eq:solveo2}. As demonstrated in Proposition \ref{prop:3} below, we can control the error due to this approximation.  As shown in Section \ref{sec:6}, this discretization approach yields effective results in practice.
\begin{proposition} \label{prop:3}
Let $(\hat{p}_{\operatorname{cDGM}}(\mathbf{x}), \hat{q}_{\operatorname{cDGM}}(\mathbf{x})) $ denote the optimal solution from \eqref{eq:solveo2}. Under Assumption \ref{assumption:3}, for any $\epsilon>0$, there exists a discrete price grid $\tilde{\mathcal{P}}$, such that 
\begin{align}
    \hat{\pi}(\mathbf{x}, \hat{p}_{\operatorname{cDGM}}(\mathbf{x}), \hat{q}_{\operatorname{cDGM}}(\mathbf{x})) \leq \hat{\pi}(\mathbf{x}, \tilde{p}_{\operatorname{cDGM}}(\mathbf{x}), \tilde{q}_{\operatorname{cDGM}}(\mathbf{x})) + \epsilon, \notag
\end{align}
for any $\bx\in\mathcal{X}$, where the grid size $|\tilde{\mathcal{P}}|\leq p_{\max}(2C+(2p_{\max}-c-s)L)/\epsilon$, where $C$ and $L$ are positive constants defined in Assumption \ref{assumption:3}.
\end{proposition}

\begin{algorithm}[htb]
\caption{Joint Inventory and Pricing Decision} \label{algo:1}
\begin{algorithmic}[1] 
    \State \textbf{Input:} The trained generator $G_{\hat{\theta}_n}(\cdot,\cdot,\cdot)$, parameter $c$ and $s$, the feature $\bx$ of the period, the number of generation $M$, discrete price set $\tilde{\mathcal{P}}=\{p_j\}_{j=1}^J$.
    \For{$p$ \textbf{in} $\tilde{\mathcal{P}}$ }  \Comment{Done with parallel computing for $\tilde{\mathcal{P}}$}
    \State Apply Algorithm \ref{algo:0} with $\bx$ and $p$ to obtain $\hat{S}(\bx,p)$ and $\hat{q}(\bx,p)$. 
    \State Estimate the expected profit by \eqref{eq:dgmest}: $\hat{\pi}(\bx,p)\leftarrow\hat{\pi}(\bx,p,\hat{q}(\bx,p))$.
    \EndFor
    \State $\tilde{p}(\bx) \leftarrow \argmax _{p\in\tilde{\mathcal{P}}} \hat{\pi}(\bx,p)$, $\tilde{q}(\bx)\leftarrow \hat{q}(\bx,\tilde{p}(\bx))$.
    \State \textbf{Output:} price $\tilde{p}(\bx)$ and order quantity $\tilde{q}(\bx)$.
\end{algorithmic}
\end{algorithm}

\subsection{Method Comparison} 
For solving problem \eqref{eq:oo1}, traditional methods often face limitations related to misspecification, feature handling, data storage, and retraining. Methods like SAA, ERM-LR, and RBE struggle with misspecification: SAA ignores features entirely, ERM-LR’s linear assumption fails to capture nonlinear demand relationships, and RBE’s additive structure leads to biased residuals. KO avoids misspecification but suffers from the curse of dimensionality, limiting its effectiveness in high-dimensional settings (e.g., $k > 5$) \citep{hastie2009elements}. Data storage and retraining requirements also vary across methods. ERM-LR, ERM-NN, and KO necessitate retraining for different price points, which requires maintaining historical data, while SAA and RBE avoid retraining by storing demand data and residuals. However, continuous data storage affects their efficiency and scalability. Our proposed cDGM-based method overcomes these challenges. Lemma \ref{lemma:1} ensures it avoids misspecification, and its neural network structure handles high-dimensional features effectively \citep{han2023deep}. With cDGMs, a single training phase suffices, and Algorithms \ref{algo:0} and \ref{algo:1} require only forward passes without historical data storage, making this approach efficient and scalable for large-scale applications. A summary of these comparisons is provided in Table \ref{tab:summary}.

\begin{table}[htbp]
    \centering
    \caption{Summary of Methods for \eqref{eq:oo1}} \label{tab:summary}
    \begin{tabular}{lcccccc}
        \toprule
        \multirow{2}{*}{Methods} & \multirow{2}{*}{\shortstack{Misspecification \\issue}}  & \multicolumn{2}{c}{Allow features} & \multirow{2}{*}{\shortstack{Retrain or \\  store data}} \\
        \cmidrule(lr){3-4}
        &  & low dim & high dim &  \\
        \midrule
        SAA        & Yes  & \xmark & \xmark & Require  \\
        ERM-LR       & Yes  & \cmark & \cmark & Require  \\
        ERM-NN      & No   & \cmark & \cmark & Require  \\
        KO         & No   & \cmark & \xmark & Require \\
        RBE        & Yes  & \cmark & \cmark & Require \\
        cDGM       & No   & \cmark & \cmark & Not Require \\
        \bottomrule
    \end{tabular}
\end{table}

For the feature-based newsvendor pricing problem \eqref{eq:oo2}, as analyzed in Section \ref{sec:3.3}, there are limited feasible methods available for comparison. RBE can approximate the profit function accurately if a linear model with independent randomness is valid, but its performance declines in more complex settings. The prescriptive approach \citep{bertsimas2020predictive} outlined in \eqref{eq:presc} is capable for handling problem \eqref{eq:oo2}. However, its choice of $w_i(\bx,p)$ relies on traditional statistic or machine learning methods, which are limited in high-dimensional contexts and lack flexibility to handle complex feature types like text or image data. Furthermore, the weighted approach requires frequent refitting or access to the entire historical data $\{d_i\}_{i=1}^n$ to handle different features  $\bx\in\mathcal{X}$, whereas our method is more efficient and portable.

\section{Theoretical Results} \label{sec:5}
In this section, we present our theoretical foundation supporting our proposed method, with focusing on the case where $(\bX,P,D)$ are continuous-type random variables. The primary theoretical question concerns the learning process of the generator $G_{\hat{\theta}}$: while an optimal solution exists by Lemma \ref{lemma:1}, it remains to be demonstrated whether training the model on historical data $S_n$ allows it to approximate this optimal solution. 

Our framework places no restrictions on the specific cDGMs used, and the strong approximation capabilities of neural networks allow various cDGMs to effectively address real-world problems. However, from a theoretical perspective, each model requires distinct analytical techniques due to differences in principles, structures, and loss functions, among other factors. While deriving theoretical guarantees for different methods are diverse and not the primary focus of this paper, one common objective across these approaches is to show that the model can approximate the true target distribution. Accordingly, we introduce a sufficient condition regarding the model’s learning process in Assumption \ref{assumption:1}. 

\begin{assumption} \label{assumption:1}
    $\lim_{n\rightarrow\infty} \mathbb{E}\left[\int_{-\infty}^{\infty}\big| F_{{G}_{\hat{\theta}}}(y|\mathbf{X},P) - f_{D|\mathbf{X},P}(y|\mathbf{X},P) \big| \, dy\right] \rightarrow 0. $
    Here, \( F_{{G}_{\hat{\theta}}} \) and \( f_{D|\mathbf{X},P} \) denote the conditional density functions of the generated demand from our trained generator \( G_{\hat{\theta}} \) and the true conditional demand \( D | \mathbf{X}, P \), respectively. The expectation is taken with respect to \( \mathbf{X} \), \( P \), and the historical data \( S_n \).
\end{assumption}
Assumption \ref{assumption:1} assumes that the conditional density function of demand samples generated by  \( G_{\hat{\theta}} \) is consistent, in the sense that it converges to the true target conditional density under the total variation distance. This assumption aligns with Theorem 4.1 in \citet{zhou2023deep}, which validates such convergence using cGANs as the generative models. Existing work on DGMs, such as by \citet{oko2023diffusion} and \citet{chang2024deep} for diffusion models, and by \citet{cherief2020convergence} and \citet{chae2023likelihood} for VAEs, explores relevant convergence properties, potentially providing valuable insights for extending our theoretical results for other cDGMs. 

\begin{assumption} \label{assumption:2}
    The marginal distribution of \((\bx, p)\) is supported on $\mathcal{X}\times\mathcal{P}$, with a continuous density function $f_{\mathbf{X},P}(\bx,p)$.
\end{assumption}

\begin{assumption} \label{assumption:3}
    Our deep neural generator \( G_{\theta}\) and the target generator $G^*$ \( : \mathcal{X} \times \mathcal{P} \times \mathbb{R}^r \rightarrow \mathbb{R}^+ \) belong to a function class \( \mathcal{G} \) such that, for any \( G \in \mathcal{G} \): (1) \( G \) is bounded, i.e., there exists a constant $C>0$, for any \( (\mathbf{x}, p, t) \in \mathcal{X} \times \mathcal{P} \times \mathbb{R}^r \), \( |G(\mathbf{X}, P, t)| < C \); and (2) \( G(\cdot, \cdot, t) \) is \( L \)-Lipschitz continuous for \( t \in \mathbb{R}^r \), i.e., there exists a constant $L>0$, for any \( (\bx_1,p_1), (\bx_2,p_2) \in \mathcal{X}\times\mathcal{P} \),
    \begin{align}
        |G(\mathbf{x}_1, p_1, t) - G(\mathbf{x}_2, p_2, t)| \leq L \|(\bx_1,p_1) - (\bx_1,p_1)\|_2; \notag
    \end{align}
    and (3) the density of $G(\bx,p,\boldsymbol{\eta})$ is upper bounded by a constant $C_1>0$ for any $(\bx,p)\in\mathcal{X}\times\mathcal{P}$ where $\boldsymbol{\eta}\sim \mathcal{N}(\mathbf{0},\mathbf{I}_r)$.
\end{assumption}
Assumption \ref{assumption:2} imposes a continuity condition for conditional density of the price and features. The boundedness and Lipschitz continuity conditions in Assumption \ref{assumption:3} imposed on the function class \( \mathcal{G} \) are relatively mild, especially compared to stricter conditions like Hölder smoothness used in other theoretical studies, \citep[e.g.,][]{schmidt2020nonparametric,jiao2023deep,han2023deep}. For Assumption \ref{assumption:3}(3), it is mild to assume a standard normal distribution retains a bounded density function after a Lipschitz continuous transformation $G$. Assumption \ref{assumption:3} also implies that the demand $D$ is bounded by $C$ and supported on a finite range, which aligns with conditions in real-world markets.

We now present our theoretical results. Let \(\xrightarrow{\mathrm{P}}\) denote convergence in probability. For a sequence of random variables indexed by \( n \) and \( M \) and random variable $A$. We say that $A_{n,M}\xrightarrow{\mathrm{P}}A$ if, for every \( \epsilon > 0 \),
\begin{align} \label{eq:mn}
\lim_{n \to \infty} \lim_{M \to \infty} \mathbb{P} \left( |A_{n,M} - A| > \epsilon \right) = 0.
\end{align}
If we choose \( M = M(n) \) such that \( M(n) \to \infty \) as \( n \to \infty \), then \eqref{eq:mn} implies that $A_{n,M(n)}\xrightarrow{\mathrm{P}}A$. We use the form in \eqref{eq:mn} to allow for an unrestricted order of \( M \). 

In our Algorithm \ref{algo:0}, our solution to problem \eqref{eq:oo1} is \( \hat{q}_{\operatorname{cDGM}}(\mathbf{x}, p) \) as a function of \( (\mathbf{x}, p)\in\mathcal{X}\times\mathcal{P} \).
\begin{theorem} \label{theorem:1}
    Under Assumptions \ref{assumption:1}–\ref{assumption:3}, for any \( (\bx,p) \in \mathcal{X}\times\mathcal{P} \), we have
    \begin{align}
        \hat{q}_{\operatorname{cDGM}}(\mathbf{x}, p) - q^{*}(\mathbf{x}, p) \xrightarrow{\mathrm{P}} 0, \notag
    \end{align}
    which implies
    \begin{align} \label{eq:them1}
        R(\bx,p):= \pi(\mathbf{x}, p, q^*(\mathbf{x}, p)) - \pi(\mathbf{x}, p, \hat{q}_{\operatorname{cDGM}}(\mathbf{x}, p))   \xrightarrow{\mathrm{P}} 0. 
    \end{align}
\end{theorem}
Theorem \ref{theorem:1} establishes that, for any given \( (\bx,p) \in \mathcal{X}\times\mathcal{P} \), the decision of our Algorithm \ref{algo:0} converges in probability to \( q^{*}(\bx,p) \) as both the sample size \( n \) and the number of generated samples \( M \) approach infinity. Additionally, it demonstrates that, for any fixed \( p \in \mathcal{P} \), the profit obtained by our decision for problem \eqref{eq:oo1} converges to the optimal profit in probability. For any given \( p \in \mathcal{P} \),  substituting the random features \( \bX \) into \( R(\bx, p) \) and taking the expectation with respect to \( \bX \) yields
\begin{align} \label{eq:excessrisk}
    \mathbb{E}_{\mathbf{X}} \left[R(\bX,p) \right] &= \mathbb{E}_{\mathbf{X},D} \left[ (c-s)(\hat{q}_{\operatorname{cDGM}}(\bX,p)-D)^+ + (p-c)(D-\hat{q}_{\operatorname{cDGM}}(\bX,p))^+ |P=p \right ] \notag \\
    &\quad -\mathbb{E}_{\mathbf{X},D} \left[ (c-s)(q^*(\bX,p)-D)^+ + (p-c)(D-q^*(\bX,p))^+ |P=p \right ].
\end{align}
This expression is similar to the form of the newsvendor excess risk as studied in \citet{han2023deep}, where $p-c$ and $c-s$ serve as the underage and overage cost, respectively. Since the demand is bounded, by \eqref{eq:them1}, the excess risk in \eqref{eq:excessrisk} converges to zero as $n,M\rightarrow\infty$. 

Next, we examine our profit estimator \( \hat{\pi}(\bx, p, q) \) in \eqref{eq:dgmest} and its role in Algorithm \ref{algo:1} during the optimization process in \eqref{eq:tildep}. We present the following theorem:
\begin{theorem} \label{theorem:2}
    Under Assumptions \ref{assumption:1}–\ref{assumption:3}, for any \( p \in \mathcal{P} \) and $q\in\mathcal{R}^+$, we have
    \begin{align}
        \hat{\pi}(\bx,p,q) \xrightarrow{P} \pi(\bx,p,q). \notag
    \end{align}
    Also, for any \( p \in \mathcal{P} \), we have
    \begin{align}
        \hat{\pi}(\mathbf{x}, p, \hat{q}_{\operatorname{cDGM}}(\mathbf{x}, p))   \xrightarrow{\mathrm{P}} \pi(\mathbf{x}, p, q^*(\mathbf{x}, p)). \notag
    \end{align}
\end{theorem}
Theorem \ref{theorem:2} establishes that \( \hat{\pi}(\bx, p, q) \) is a consistent estimator of the true conditional profit function \( \pi(\bx, p, q) \). Additionally, it shows that, for a fixed price, the estimated profit under our order quantity decision converges to the optimal true profit at that price. This result provides a theoretical foundation for the pricing decision in Algorithm \ref{algo:1}. 

We now present Theorem \ref{theorem:3}, which provides a theoretical guarantee for the joint decision derived from Algorithm \ref{algo:1} over any discrete price set \( \mathcal{P} = \{p_j\}_{j=1}^J \).

\begin{theorem}  \label{theorem:3}
    For any discrete price set \( \tilde{\mathcal{P}} = \{p_j\}_{j=1}^J \subset \mathcal{P} \), under Assumptions \ref{assumption:1}–\ref{assumption:3}, the decisions obtained by our Algorithm \ref{algo:1}, 
    \begin{align}
        (\tilde{p}_{\operatorname{cDGM}}(\mathbf{x}), \tilde{q}_{\operatorname{cDGM}}(\mathbf{x})) &= \underset{p \in \tilde{\mathcal{P}}, q \in \mathbb{R}^+}{\argmax} \ \hat{\pi}(\mathbf{x}, p, q), \notag
    \end{align}
    are asymptotically optimal within \( \tilde{\mathcal{P}} \) in the sense that 
    \begin{align} \label{eq:them3}
    \max  _{p\in\tilde{\mathcal{P}},q\in\mathbb{R}^+} \pi(\bx,p,q)  - \pi(\mathbf{x}, \tilde{p}_{\operatorname{cDGM}}(\mathbf{x}), \tilde{q}_{\operatorname{cDGM}}(\mathbf{x})) \xrightarrow{\mathrm{P}} 0. 
    \end{align}
\end{theorem}
Theorem \ref{theorem:3} demonstrates that the profit achieved by our joint decision converges in probability to the optimal true profit within any discrete price set $\tilde{\mathcal{P}}$. Note that the term $\tilde{p}_{\operatorname{cDGM}}(\mathbf{x})$ is directly plugged into $\pi(\cdot,\cdot,\cdot)$ in \eqref{eq:them3}, whose randomness does not affect the conditional expectation related to the price taken inside $\pi(\cdot,\cdot,\cdot)$. Based on Theorem \ref{theorem:3}, we further have the following corollary.
\begin{corollary} \label{coro:1}
    Under Assumptions \ref{assumption:1}–\ref{assumption:3}, for any tolerance level $\epsilon>0$, there exists a price grid $\tilde{\mathcal{P}}$ with size $|\tilde{\mathcal{P}}|\leq p_{\max}(2C+(2p_{\max}-c-s)L)/\epsilon$, such that the profit obtained by our Algorithm \ref{algo:1} converges in probability to a value at least $\pi(\bx,p_{\operatorname{opt}}(\bx),q_{\operatorname{opt}}(\bx)) -\epsilon$. 
\end{corollary}
Corollary \ref{coro:1} establishes that the discretization method employed in Algorithm \ref{algo:1} can yield profits arbitrarily close to those of the true optimal solution over the continuous set $\mathcal{P}$.

In summary, we provide a rigorous theoretical foundation for our cDGM-based approach. The convergence results established in Theorems \ref{theorem:1}–\ref{theorem:3} and Corollary \ref{coro:1} demonstrate our method approximates the true optimal solutions under realistic conditions. To further validate the effectiveness of our proposed methods, we proceed with experiments in Section \ref{sec:6} and Section \ref{sec:7}.

\section{Simulation Study} \label{sec:6}
In this section,  we conduct a simulation study to evaluate the performance of our proposed methods. The simulations are based on the following data-generating processes (DGPs) for $D|(\bX=\bx,P=p)$:
\begin{itemize}
    \item[(a)] $ = 100 - 20p + \bx^\top\boldsymbol{\beta} + \epsilon$, $\quad \epsilon\sim \mathcal{N}(0,5)$
    \item[(b)] $ = 100 - 20p + 4\sin(2x_1) + 3x_2x_3 + \epsilon$, $\quad \epsilon\sim \mathcal{N}(0,5)$
    \item[(c)] $ = 130(4p-6)^{-1.3}\epsilon + \bx^\top\boldsymbol{\beta} $, $\quad \log\epsilon\sim \mathcal{N}(0,0.5)$
    \item[(d)]  $= 40(4-p)^{\sin(3g(\mathbf{x}))+1.01} + \epsilon$, $\quad \epsilon \sim \mathcal{N}(0,4)$
\end{itemize}
In these DGPs, the 5-dimensional feature vector $\bx=(x_1,x_2,...,x_5)^{\top}$ follows a multivariate normal distribution, $\bx\sim \mathcal{N}(\mathbf{0},\Omega)$, where $\Omega_{ii}=1$ and $\Omega_{ij}=0.5$ for $i\neq j$. The coefficients $\boldsymbol{\beta}\sim \mathcal{N}(\mathbf{0},2\mathbf{I}_5)$, and $g(\bx)=\mathbf{x}^\top\boldsymbol{\beta_e}/\sqrt{15}$, where $\boldsymbol{\beta}_{e}=(1,...,1)^\top\in\mathbb{R}^5$. The formulation of $g(\bx)$ is to make sure $g(\bx)\sim \mathcal{N}(0,1)$. The retail cost is set at $c=1.0$ and salvage value at $s=0.5$. Historical prices follows uniform distribution on $\mathcal{P}$ where $\mathcal{P}\subset [c,4]$ is either a discrete set or a continuous interval. Demand values across (a)-(d) are truncated to the range $[0,200]$.

In DGPs (a) and (b), the influence of features on demand is relatively small, while in DGPs (c) and (d), the features have larger influence on the demand's distribution. These DGPs cover a range of complexities in demand from simple to complex, providing a robust framework to validate the proposed methods.

\subsection{Discrete Price Set} \label{sec:6.1}
We start with a scenario where the price set $\mathcal{P}$ is discrete. In the historical observations $S_n=\{(\bx_i,p_i,d_i)\}_{i=1}^n$ with $n=2000$, $\mathcal{P}$ is a discrete set of 21 prices within $[2,4]$ for DGPs (a)-(c) and $[1,4]$ for DGP (d).

\subsubsection{Inventory Decision (O-A)} \label{sec:6.1.1} 
First, we focus on inventory decisions across different methods. For each $p\in\mathcal{P}$, we define $S_{n,p}=\{(\bx_i,p_i,d_i):p_i=p\}$ as the subset of training data where the price is fixed at $p$. SAA uses empirical quantile of $S_{n,p}$ independently for each price, while ERM-based methods and KO train/fit models based on $S_n$ with quantile level $\rho(p)$. In contrast, our cDGM-based method requires only one training on $S_n$ and can be applied to every $p\in\mathcal{P}$ without retraining.

For testing, we consider a test set $S_{\operatorname{test}}=\bigcup_{p\in\mathcal{P}} S_{\operatorname{test},p}$, where $S_{\operatorname{test},p} = \{(\bx_{i,p}^{\operatorname{test}},p,d_{i,p}^{\operatorname{test}})\}_{i=1}^{n_{\operatorname{test}}}$ with $n_{\operatorname{test}}=1000$ for each $p\in \mathcal{P}$. Each method makes an inventory decision, and we calculate the mean profit for each decision over the test samples. To evaluate performance, for each $p\in\mathcal{P}$, we calculate the average difference between the profit achieved by each method $\hat{q}(\cdot,p)$ and that of the oracle method $q^*(\cdot,p)$, given by
\begin{align} 
    L_{\operatorname{test}}(p) = \frac{1}{n_{\operatorname{test}}}\sum_{i=1}^{n_{\operatorname{test}}}  \Pi(d_{i,p}^{\operatorname{test}},p,q^*(\bx_{i,p}^{\operatorname{test}},p))-\Pi(d_{i,p}^{\operatorname{test}},p,\hat{q}(\bx_{i,p}^{\operatorname{test}},p)),  \notag
\end{align}
which is an estimate of excess risk \eqref{eq:excessrisk} of $\hat{q}(\cdot,p)$. By Theorem \ref{theorem:1}, our method should perform well and outperform other methods that suffer from model misspecification theoretically. 
We also calculate the average profit difference across all prices in $\mathcal{P}$: $\sum_{p\in\mathcal{P}}L_{\operatorname{test}}(p)/|\mathcal{P}|$, shown in Table \ref{tab:0.5}.

\begin{table}[htbp]
\centering
\begin{tabular}{lcccccc}
\toprule
DGP & SAA & RBE & ERM-LR & ERM-NN & KO & cDGM \\ 
\midrule
(a) & 6.51 (0.49) & \textbf{\textcolor{red}{0.00}} (0.00) & \underline{\textcolor{blue}{0.01}} (0.01) & 0.33 (0.04) & 2.54 (0.11) & 0.04 (0.03) \\ 
(b) & 6.42 (0.09) & 1.25 (0.06) & 1.25 (0.06) & \textbf{\textcolor{red}{0.35}} (0.04) & 1.10 (0.06) & \underline{\textcolor{blue}{0.61}} (0.04) \\ 
(c) & 11.25 (0.50) & 2.32 (0.19) & 2.32 (0.19) & \underline{\textcolor{blue}{1.21}} (0.19) & 7.96 (0.28) & \textbf{\textcolor{red}{0.95}} (0.23) \\ 
(d) & 32.02 (0.94) & 15.67 (0.21) & 15.67 (0.27) & \underline{\textcolor{blue}{2.78}} (0.21) & 14.55 (0.21) & \textbf{\textcolor{red}{2.64}} (0.34) \\ 
\bottomrule
\end{tabular}
\caption{Inventory Decision Results (Discrete): Comparison of Profit Difference with the Oracle Inventory}
\label{tab:0.5}
\end{table}

Each experiment is repeated 50 times. In simpler cases, such as DGP (a), methods like RBE and ERM-LR perform similarly to the oracle, as their structures align well with the true model. However, in scenarios involving nonlinear dependencies or complex feature interactions, as seen in DGP (b), both our method and ERM-NN show a clear advantage, as summarized in Table \ref{tab:0.5}. Visualizations of $L_{\operatorname{test}}(p)$ for DGPs (a) and (b) can be found in the Appendix B. Figures \ref{fig:1} and \ref{fig:2} display $L_{\operatorname{test}}(p)$ across prices for DGPs (c) and (d) with shaded areas representing variability across runs. Results indicate that our method achieves consistently small profit differences from the oracle across most prices compared to other methods. ERM-NN also performs well, leveraging neural networks for complex relationships. 

In summary, our cDGM-based method performs consistently well across a range of DGPs, demonstrating its strengths particularly in complex scenarios involving nonlinear structures and feature-dependent randomness. While our method is not expected to significantly outperform ERM-NN for the inventory-only decision problem in \eqref{eq:oo1}, ERM-NN is not feasible to handle joint inventory-pricing decisions, which we explore next in \eqref{eq:oo2}.

\begin{figure}[htbp]
    \centering
    \begin{minipage}{0.48\textwidth}
        \centering
        \includegraphics[width=\linewidth]{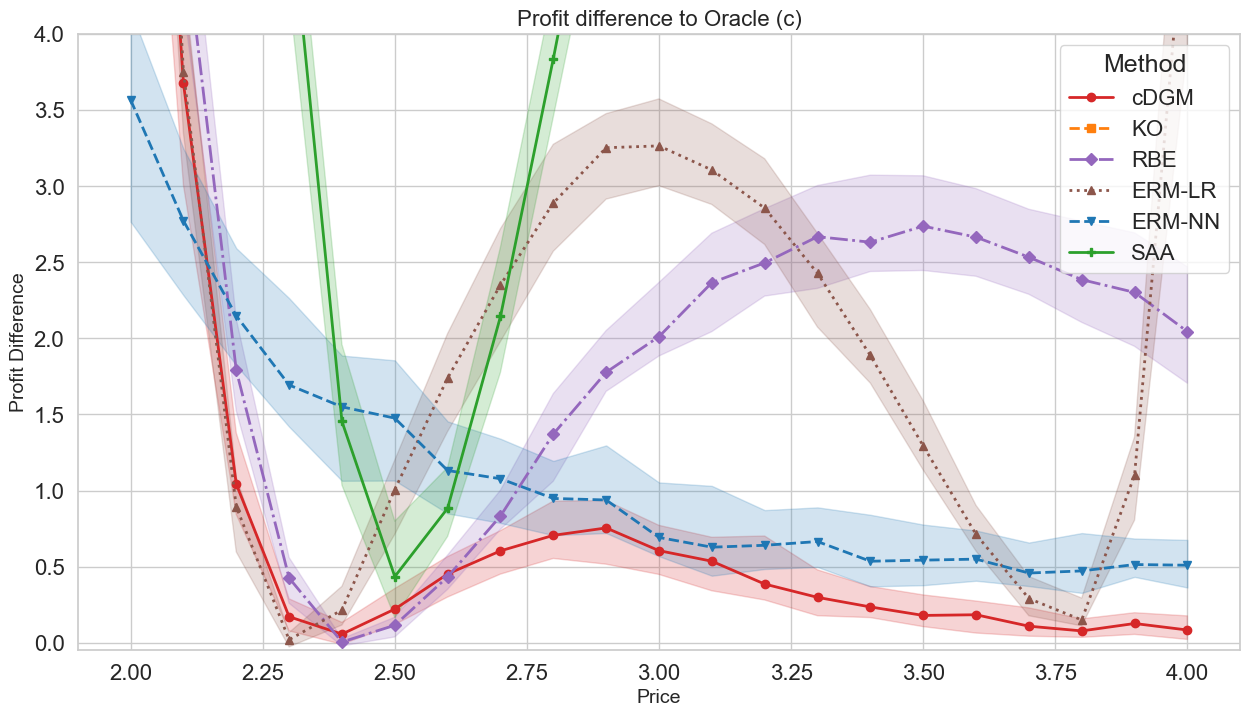}
        \caption{Inventory Results for DGP (c)}
        \label{fig:1}
    \end{minipage}%
    \hfill
    \begin{minipage}{0.48\textwidth}
        \centering
        \includegraphics[width=\linewidth]{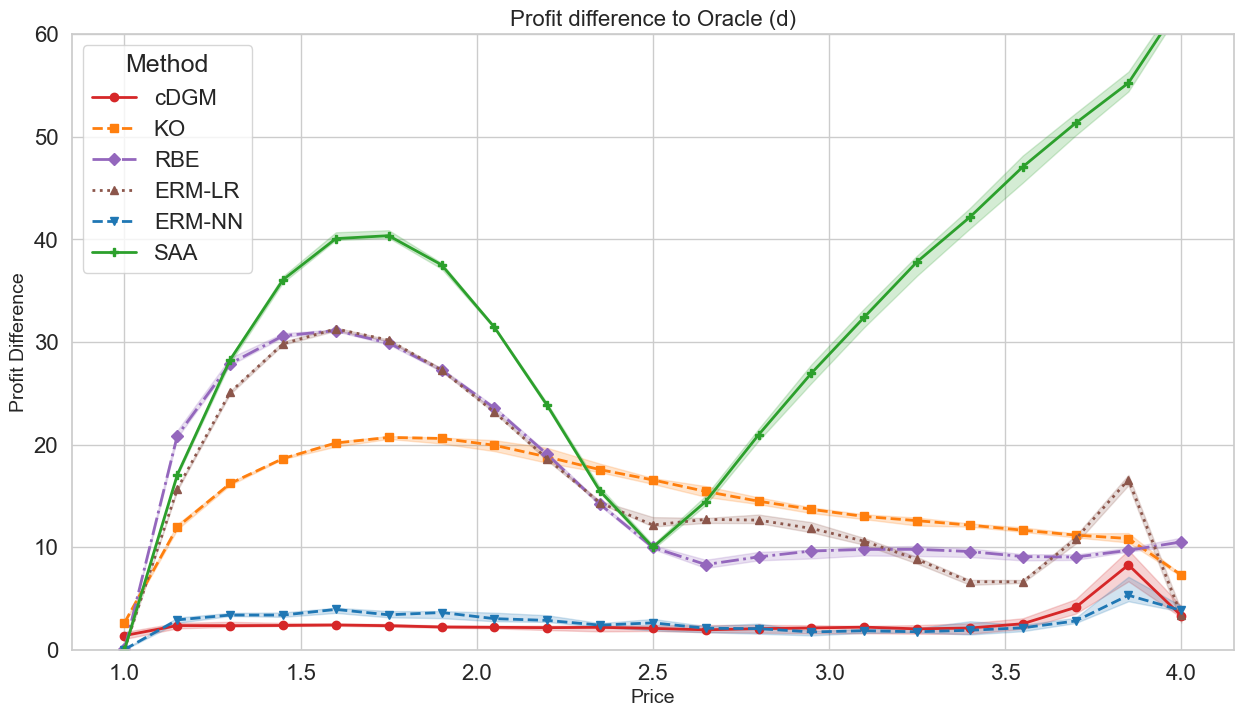}
        \caption{Inventory Results for (d)}
        \label{fig:2}
    \end{minipage}
\end{figure}

\subsubsection{Joint Inventory and Pricing Decisions (O-B)}  \label{sec:6.1.2}
We conduct experiments for joint inventory and pricing decisions in the discrete prices setting. In this scenario, we still have historic data $S_n=\{(\bx_i,p_i,d_i) \}$ for training, but during the test phase, only the features are observed for each period; no price is given as it was in the inventory-only experiments in Section \ref{sec:6.1.1}. Our task is to make simultaneous inventory and pricing decisions for a total of $n_{\operatorname{test}}=5000$ random samples $\{\bx_i^{\operatorname{test}}\}_{i=1}^{n_{\operatorname{test}}}$. For each test sample $\bx_i^{\operatorname{test}}$ in the test set, we decide both the price and the inventory level based on each method. Once the price is set to $p$, the demand $d_{i}^{\operatorname{test}}$ is determined by sampling from $D|(\bX=\bx_i^{\operatorname{test}},P=p)$. 

For comparison, we include the kernel methods as representative of the prescriptive approach introduced by \citet{bertsimas2020predictive}. This leaves three feasible methods for comparison: Kernel, RBE, and our cDGM-based method. Additionally, we use $S_n$ to construct an ``incorrect" empirical profit function estimate under the SAA inventory decision, disregarding the effect of price changes on demand:
\begin{align}
    \hat{\pi}_{S_n}(\bx,p,\hat{q}_{\operatorname{SAA}}(\bx,p)) = \frac{1}{n}\sum_{i=1}^n\Pi(d_i,p,\hat{q}_{\operatorname{SAA}}(\bx,p)), \notag
\end{align}
and choose the optimal price accordingly to evaluate its performance.

To evaluate performance, we calculate the average profit obtained by each method's decisions $(\tilde{p}(\cdot),\tilde{q}(\cdot))$ over the test set:
\begin{align} \label{eq:ave_profit}
        \frac{1}{n_{\operatorname{test}}}\sum_{i=1}^{n_{\operatorname{test}}} \Pi(d_{i}^{\operatorname{test}},\tilde{p}(\bx_{i}^{\operatorname{test}}) ,\tilde{q}(\bx_{i}^{\operatorname{test}})).
\end{align}
We repeat each simulation $50$ times, reporting the average profit and standard deviation across runs in Table \ref{tab:1}. The results demonstrate that our cDGM-based method consistently outperforms other methods across all settings and exhibits the lowest standard deviation in most DGPs, showcasing its robustness and effectiveness in optimizing joint inventory and pricing decisions to maximize profit. Notably, SAA consistently selects the maximum price, and its poor performance underscores the significant profit loss that can result from using an unsuitable modeling approach.

\begin{table}[htbp]
\centering
\begin{tabular}{lcccc}
\toprule
DGP & SAA & RBE & Kernel & cDGM \\ 
\midrule
(a) & 41.99 (0.51) & \underline{\textcolor{blue}{76.34}} (0.42) & 72.30 (0.45) & \textbf{\textcolor{red}{76.65}} (0.42) \\ 
(b) & 46.56 (0.55) & \underline{\textcolor{blue}{77.45}} (0.45) & 77.04 (0.41) & \textbf{\textcolor{red}{78.86}} (0.26) \\ 
(c) & 47.87 (3.85) & 67.84 (3.93) & \underline{\textcolor{blue}{69.96}} (3.01) & \textbf{\textcolor{red}{78.26}} (5.91) \\ 
(d) & -61.07 (2.60) & \underline{\textcolor{blue}{77.96}} (1.40) & 73.71 (1.45) & \textbf{\textcolor{red}{103.54}} (0.55) \\ 
\bottomrule
\end{tabular}
\caption{Joint Decision Results (Discrete): Profit Comparison}
\label{tab:1}
\end{table}

\subsection{Continuous Price Set} \label{sec:6.2}
We now extend our experiments to the setting where the price set $\mathcal{P}$ is continuous. Similar to the discrete case in Section \ref{sec:6.1}, we utilize historical data $S_n=\{(\bx_i,p_i,d_i)\}_{i=1}^n$ with $n=2000$ observations, but here, prices $p_i$ are drawn uniformly from the interval $[2,4]$ for DGPs (a)-(c) and $[1,4]$ for DGP (d).

\subsubsection{Inventory Decision (O-A)} \label{sec:6.2.1}
For inventory decisions with continuous prices, we adjust our evaluation to match this setting by sampling test prices uniformly over the interval. This forms a test set $S_{\operatorname{test}}=\{(\bx_i^{\operatorname{test}},p_i^{\operatorname{test}},d_i^{\operatorname{test}})\}$ of total size $n_{\operatorname{test}}=5000$.

In implementing baseline methods for continuous prices, several adjustments are necessary. For the SAA method, we operate on $S_n$. ERM-based methods present additional challenges: each sample $(\bx_i^{\operatorname{test}},p_i^{\operatorname{test}})$ in the test set  would ideally require a model with a unique quantile level $(p_i^{\operatorname{test}}-c)/(p_i^{\operatorname{test}}-s)$. To address this, we train ERM models at certain fixed quantile levels and select the most appropriate one for each test sample, maintaining efficiency while possible biases could be introduced by this approach. 

We compute the average profit for each method $\hat{q}(\cdot,\cdot)$ over $S_{\operatorname{test}}$
\begin{align} 
        \frac{1}{n_{\operatorname{test}}}\sum_{i=1}^{n_{\operatorname{test}}} \Pi(d_{i}^{\operatorname{test}}, p_i^{\operatorname{test}},\hat{q}(\bx_{i}^{\operatorname{test}},p_i^{\operatorname{test}})). \notag
\end{align}
The results are presented in Table \ref{tab:3}. Our method continues to perform consistently well across different DGPs. Notably, for ERM-based methods, the impact of quantile misspecification is less pronounced in simpler DGPs (a)-(c), where the continuous price setting has less samples with boundary price, resulting in better prediction accuracy and profits. However, in the more complex DGP (d), performance can decline due to quantile mismatch.
\begin{table}[htbp]
\centering
\begin{tabular}{lcccccc}
\toprule
DGP & SAA & RBE & ERM-LR & ERM-NN & KO & cDGM \\ 
\midrule
(a) & 6.19 (0.50) & \textbf{\textcolor{red}{0.01}} (0.01) & 0.07 (0.03) & 0.40 (0.12) & 2.53 (0.15) & \underline{\textcolor{blue}{0.04}} (0.03) \\ 
(b) & 6.12 (0.16) & 1.27 (0.10) & 1.34 (0.10) & \textbf{\textcolor{red}{0.43}} (0.13) & 1.10 (0.10) & \underline{\textcolor{blue}{0.62}} (0.07) \\ 
(c) & 9.64 (0.53) & 1.98 (0.26) & 1.76 (0.30) & \underline{\textcolor{blue}{1.50}} (0.45) & 7.83 (0.55) & \textbf{\textcolor{red}{0.87}} (0.26) \\ 
(d) & 31.61 (0.97) & 16.57 (0.32) & 19.97 (0.55) & \underline{\textcolor{blue}{3.10}} (0.37) & 14.00 (0.28) & \textbf{\textcolor{red}{2.58}} (0.54) \\ 
\bottomrule
\end{tabular}
\caption{Inventory Decision Results (Continous): Comparison of Profit Difference with the Oracle Inventory} 
\label{tab:3}
\end{table}

\subsubsection{Joint Inventory and Pricing Decisions (O-B)} \label{sec:6.2.2}
For joint inventory and pricing decisions with a continuous price set, we evaluate the average profit obtained by each method’s decisions as in \eqref{eq:ave_profit}. The same set of baseline methods is compared, and the results are presented in Table \ref{tab:4}. Our method consistently outperforms all other baselines across all DGPs, with especially large performance gaps in complex DGPs (c) and (d). 
\begin{table}[htbp]
\centering
\begin{tabular}{lcccc}
\toprule
DGP & SAA & RBE & Kernel & cDGM  \\ 
\midrule
(a) & 44.89 (0.63) & \underline{\textcolor{blue}{76.38}} (0.46) & 72.49 (0.50) & \textbf{\textcolor{red}{76.77}} (0.43) \\ 
(b) & 49.41 (0.51) & \underline{\textcolor{blue}{77.45}} (0.40) & 76.99 (0.41) & \textbf{\textcolor{red}{78.92}} (0.35) \\ 
(c) & 50.86 (3.84) & 68.52 (3.64) & \underline{\textcolor{blue}{70.20}} (2.91) & \textbf{\textcolor{red}{79.00}} (5.50) \\ 
(d) & -30.46 (2.07) & 78.06 (1.78) & \underline{\textcolor{blue}{78.95}} (0.68) & \textbf{\textcolor{red}{103.54}} (0.57) \\ 
\bottomrule
\end{tabular}
\caption{Joint Decision Results (Continous): Profit Comparison}
\label{tab:4}
\end{table}

\subsection{Simulation with Textual Features} \label{sec:6.3}
In this simulation, we evaluate the effectiveness of our method when incorporating textual features into demand learning. We consider a feature $\bx$ that consists of textual information corresponding to an unobserved score, $x_{\text{score}} \in \mathbb{R}$, which influences the demand distribution in the following linear form:
\begin{align}
(e)\ D|(\bX=\bx,P=p) = 40 + 10x_{\text{score}} - 10p + \epsilon, \notag
\end{align}
where $\epsilon\sim \mathcal{N}(0,10)$. The textual feature $\bx$ includes words that represent different scores in the range $\{1,2,3,4,5\}$, and the score $x_{\text{score}}$  is calculated as the average score of the words in each description. This score is mapped from a predefined word-to-score dictionary. For example, if $\bx = \text{``excellent, recommended"}$, where ``excellent” maps to 5 and ``recommended” maps to 4, then $x_{\text{score}} = (5+4)/2 = 4.5$. If $\bx$ is an empty text, then $x_{\text{score}} =3$. Our cDGM-based model is adjusted to process such textual input by incorporating a word embedding layer in the neural network structure (see Appendix B for implementation details).

To assess the effectiveness of incorporating textual features, we evaluate the model’s performance in the joint decision scenario. We refer to this modified model as cDGM-Text. For comparison, other methods are tested without the textual feature. Results are shown in Table \ref{tab:text}.
\begin{table}[htbp]
\centering
\begin{tabular}{lccccc}
\toprule
DGP & SAA & RBE & Kernel & cDGM & cDGM-Text \\ 
\midrule
(e) & 73.2 (4.0) & \underline{\textcolor{blue}{79.9}} (4.4) & 79.6 (4.1) & 79.3 (3.7) & \textbf{\textcolor{red}{86.0}} (3.8) \\ 
\bottomrule
\end{tabular}
\caption{Joint Decision Results (Textual): Profit Comparison}
\label{tab:text}
\end{table}
The results indicate that cDGM-Text outperforms all other methods, including the cDGM model without textual features, while other methods, except SAA, perform comparably in this DGP with linear structure. Although this setup is simplified, real-world textual features may involve more complex sentences and even paragraphs. Given the success of deep learning in natural language processing (NLP), our findings suggest that deep learning models like cDGM-Text can better leverage complex textual information, demonstrating the potential advantages of using DNNs for decision-making scenarios involving rich data types such as text and images.

\section{Real Data}  \label{sec:7}
We use the Kaggle Food Demand Forecasting dataset\footnote{\url{https://www.kaggle.com/datasets/kannanaikkal/food-demand-forecasting}} to evaluate our proposed methods. This dataset provides historical weekly demand data for meal deliveries across various fulfillment centers, including features such as meal ID, center ID, week, number of orders, base price, checkout price, and promotion flags. We treat the number of orders as the demand $d$, the checkout price as the price $p$, and select several additional attributes as features, including demand data from the previous two weeks.

\begin{figure}
    \centering
    \includegraphics[width=0.95\linewidth]{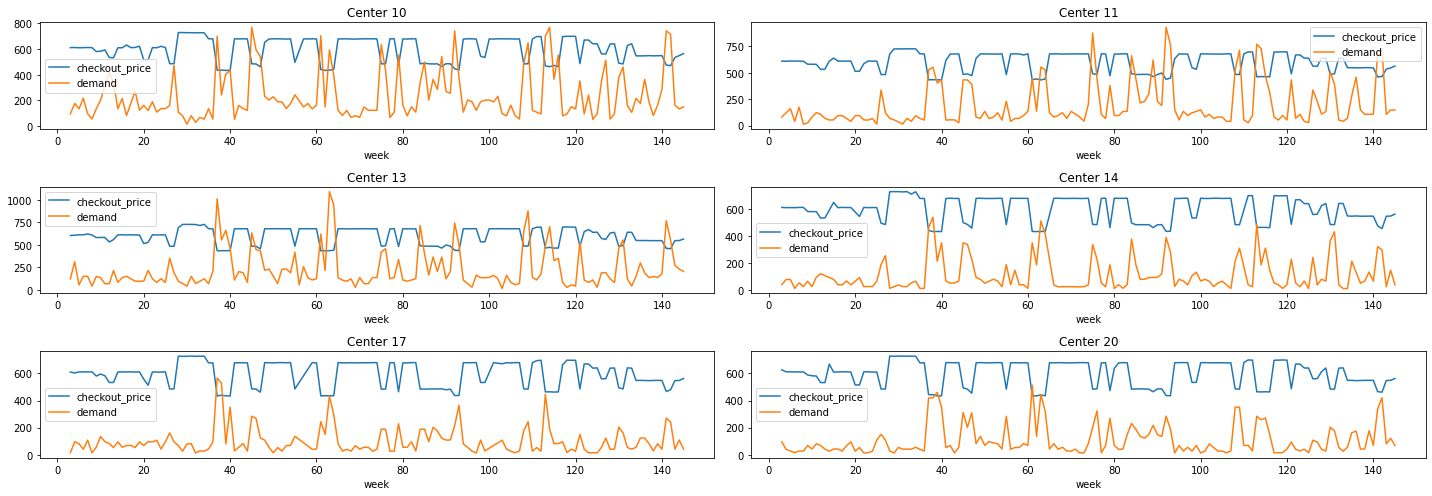}
    \caption{Visualization of price versus demand for meal ID 1558 in different centers}
    \label{fig:kaggle}
\end{figure}

Our experiments is conducted on the training subset\footnote{Only the training data is used, as the test set lacks true demand values.} with 450,730 records covering 51 meals across 77 centers over 145 weeks. For this study, we selected meals that exhibit predictable demand patterns suited to data-driven modeling. Figure \ref{fig:kaggle} shows the demand and price trends for meal ID 1558, demonstrating noticeable price elasticity and demand variations across centers, which underscores the value of incorporating features like center ID.
While we can set prices for meal records using Algorithm \ref{algo:1}, evaluating joint pricing and inventory decisions is challenging due to the inability to observe demand at alternative prices. Instead, we focus on inventory decisions for a given price (Algorithm \ref{algo:0}), similar to the setup in Section \ref{sec:6.2.1}. For ERM-based methods, we use the average price per meal and a corresponding quantile level to fit decision functions.

\begin{figure}[htbp]
    \centering
    \begin{minipage}{0.3\textwidth}
        \centering
        \includegraphics[width=\linewidth]{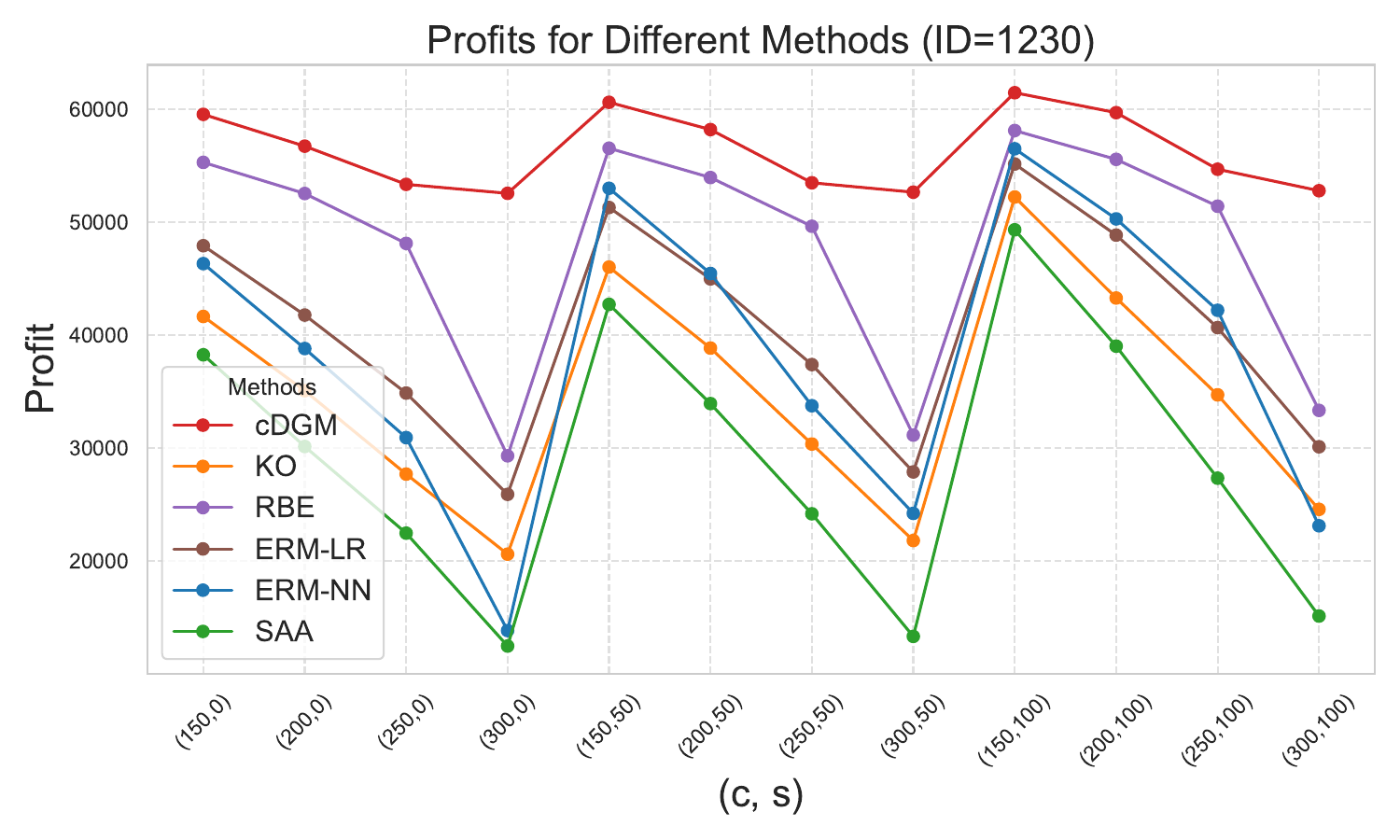}
    \end{minipage}%
    \hspace{0.02\textwidth}
    \begin{minipage}{0.32\textwidth}
        \centering
        \includegraphics[width=\linewidth]{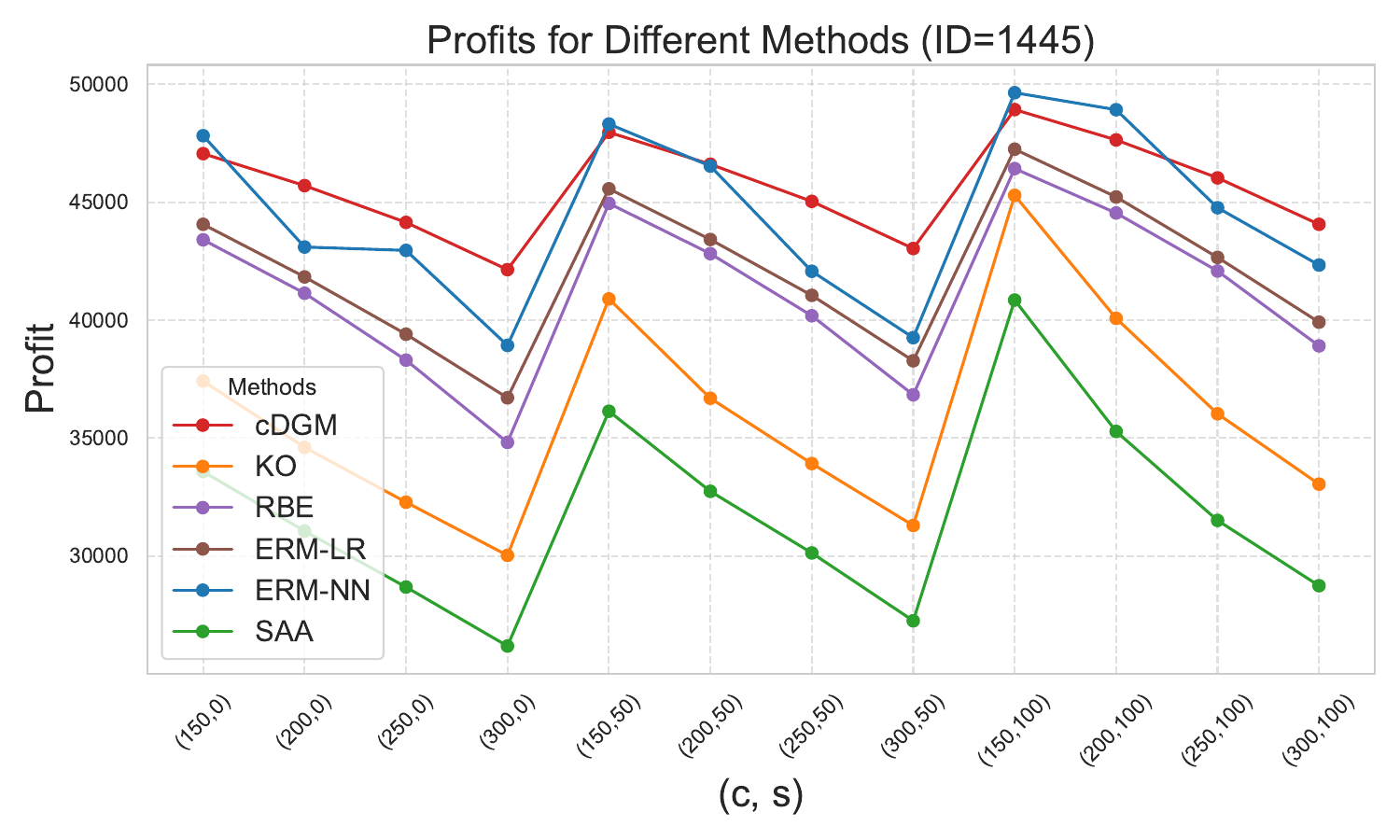}
    \end{minipage}%
    \hspace{0.02\textwidth}
    \begin{minipage}{0.3\textwidth}
        \centering
        \includegraphics[width=\linewidth]{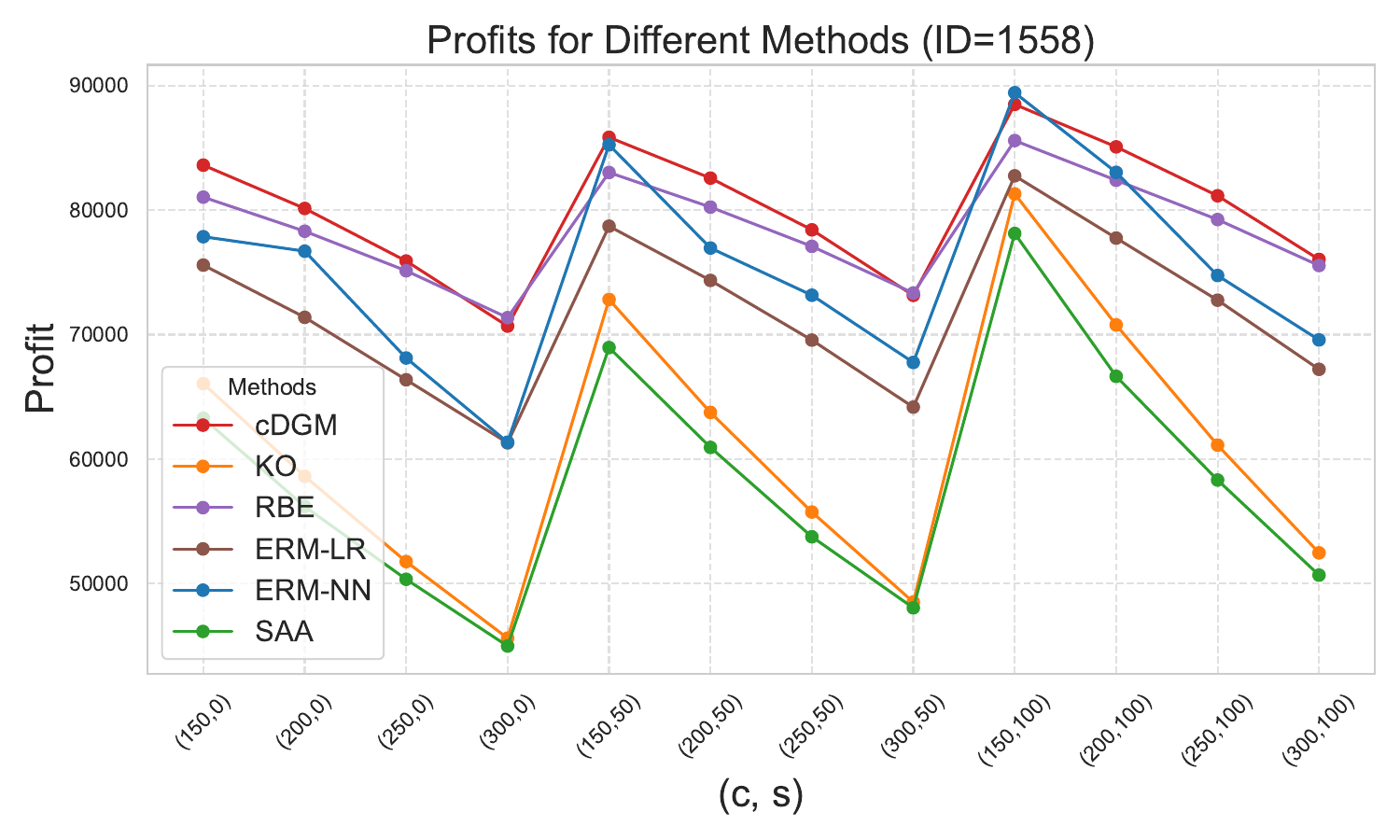}
    \end{minipage}
    \hspace{0.02\textwidth}
    \begin{minipage}{0.3\textwidth}
        \centering
        \includegraphics[width=\linewidth]{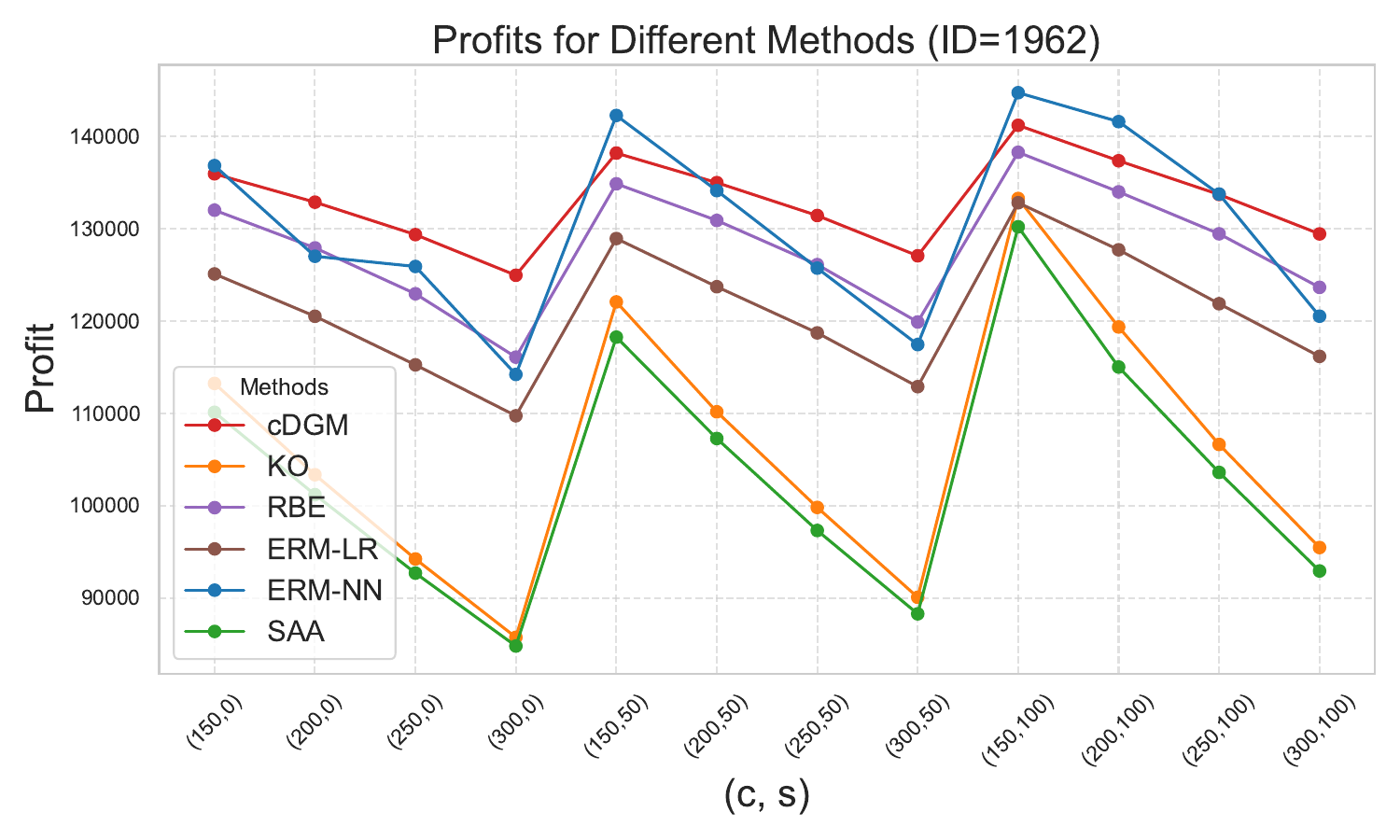}
    \end{minipage}%
    \hspace{0.02\textwidth}
    \begin{minipage}{0.3\textwidth}
        \centering
        \includegraphics[width=\linewidth]{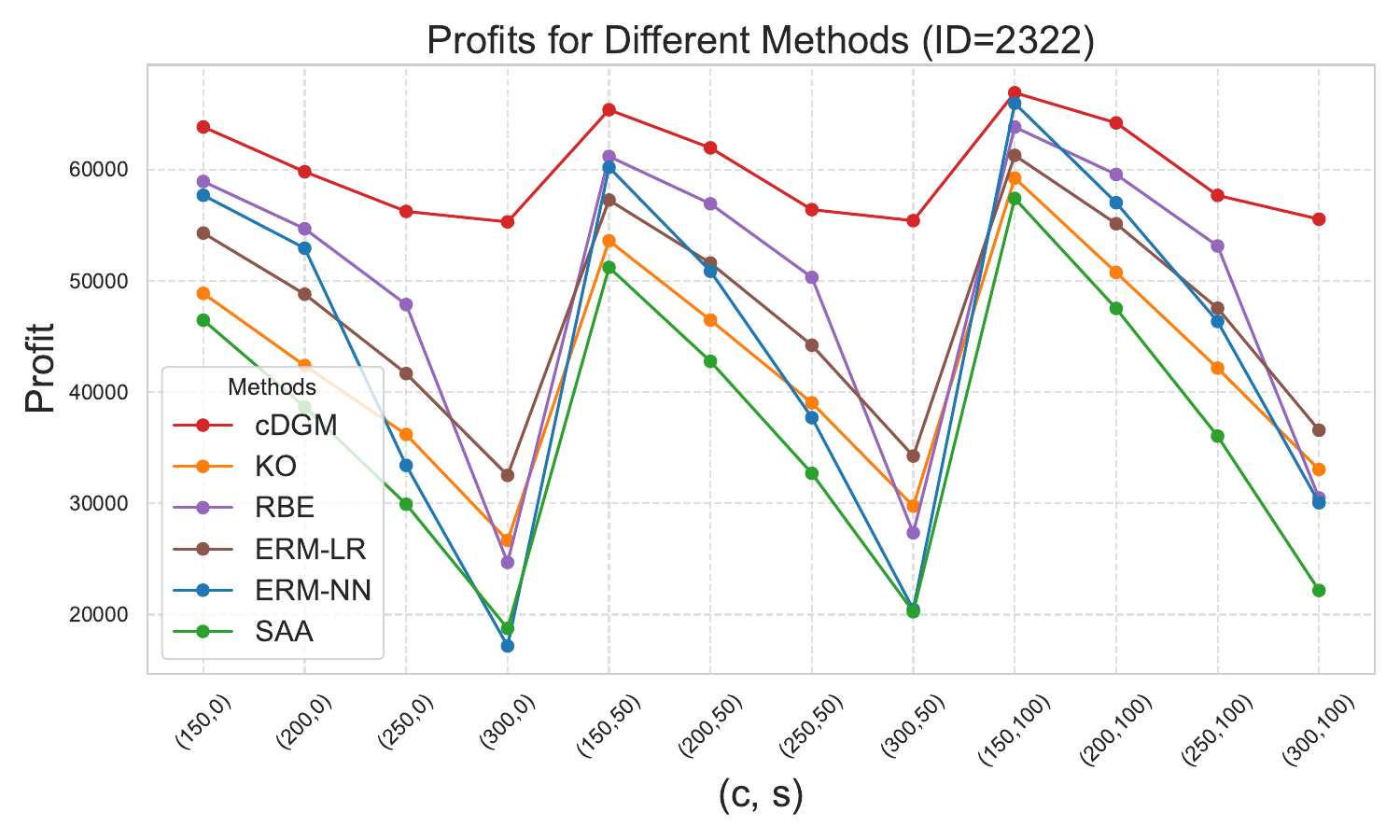}
    \end{minipage}%
    \hspace{0.02\textwidth}
    \begin{minipage}{0.3\textwidth}
        \centering
        \includegraphics[width=\linewidth]{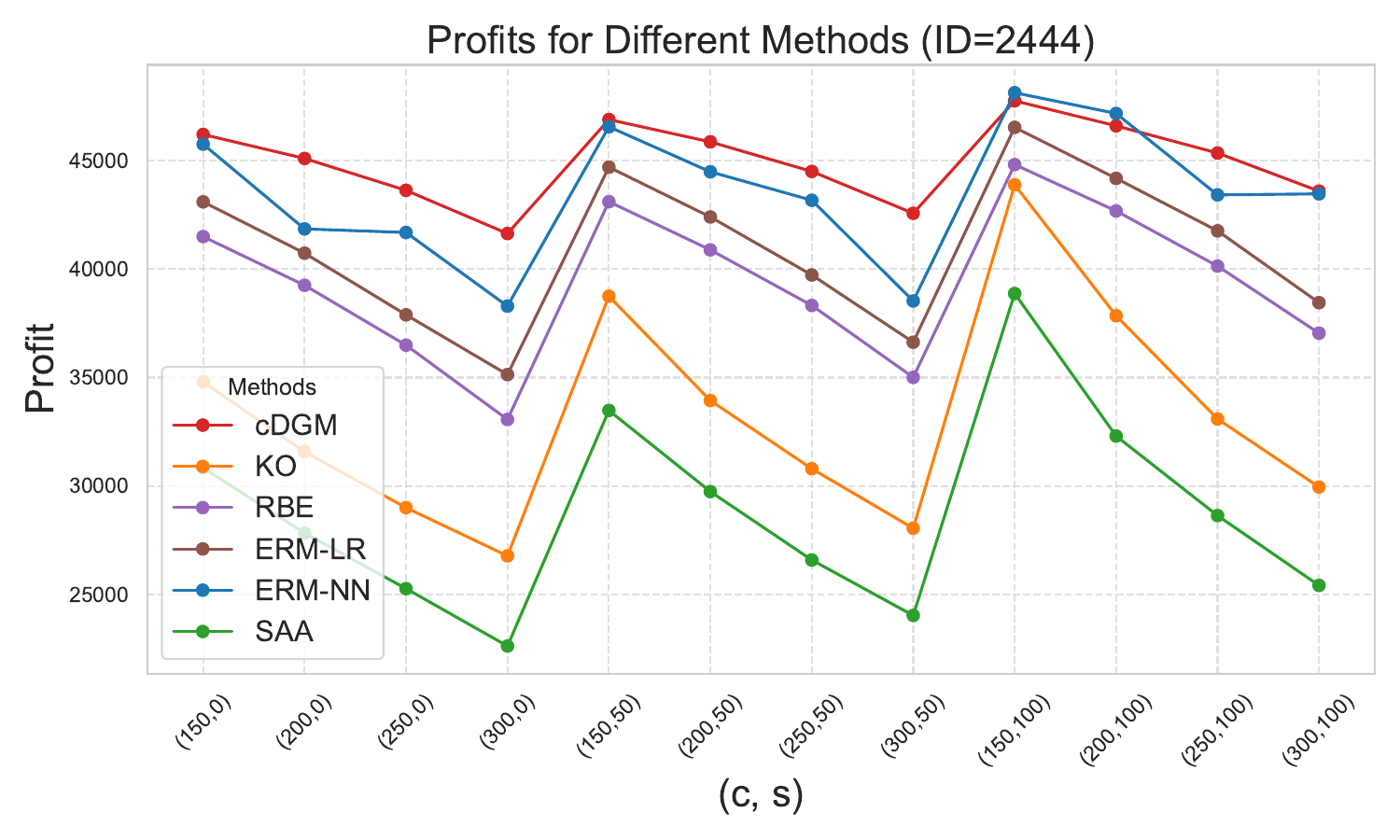}
    \end{minipage}
    \hspace{0.02\textwidth}
    \begin{minipage}{0.3\textwidth}
        \centering
        \includegraphics[width=\linewidth]{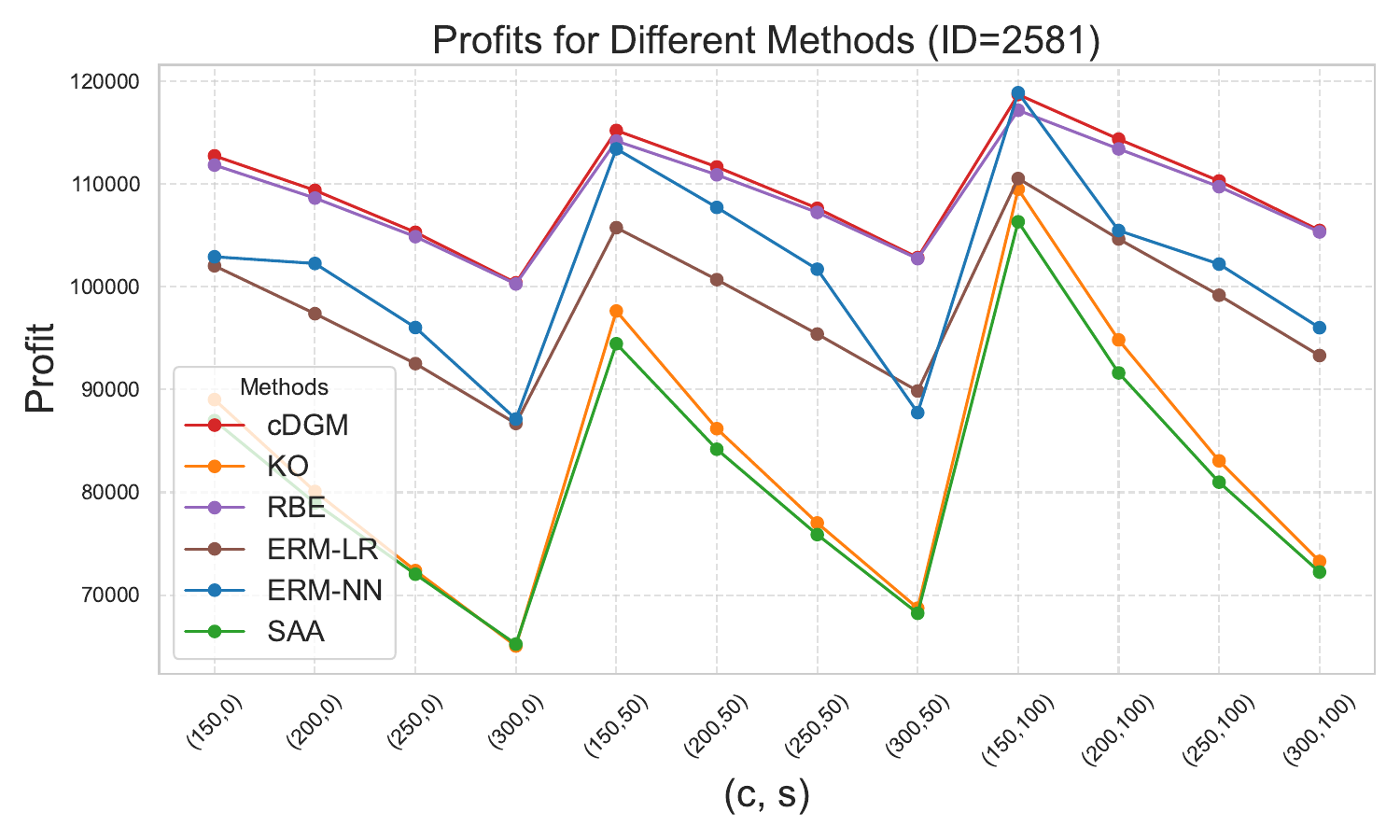}
    \end{minipage}%
    \hspace{0.02\textwidth}
    \begin{minipage}{0.3\textwidth}
        \centering
        \includegraphics[width=\linewidth]{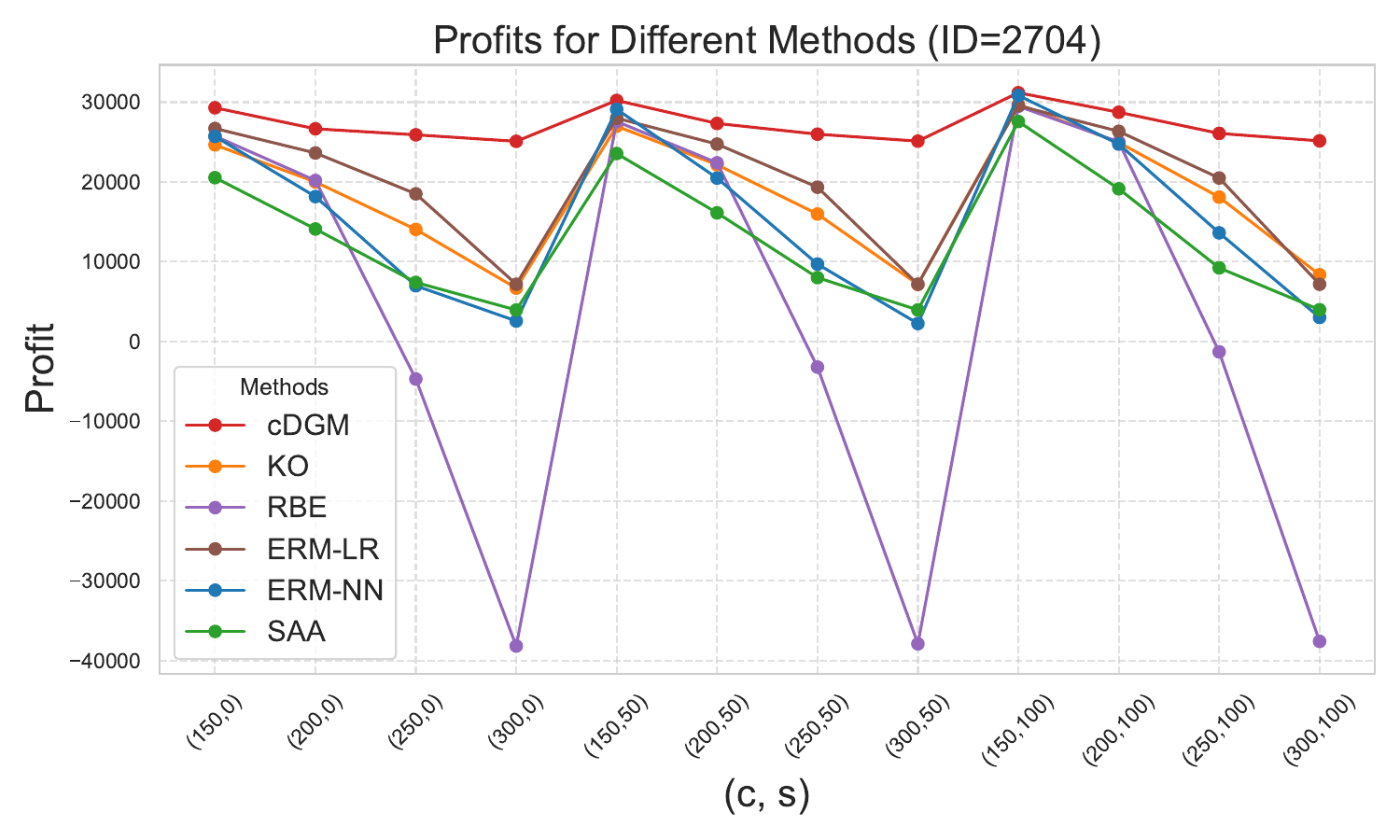}
    \end{minipage}%
    \hspace{0.02\textwidth}
    \begin{minipage}{0.3\textwidth}
        \centering
        \includegraphics[width=\linewidth]{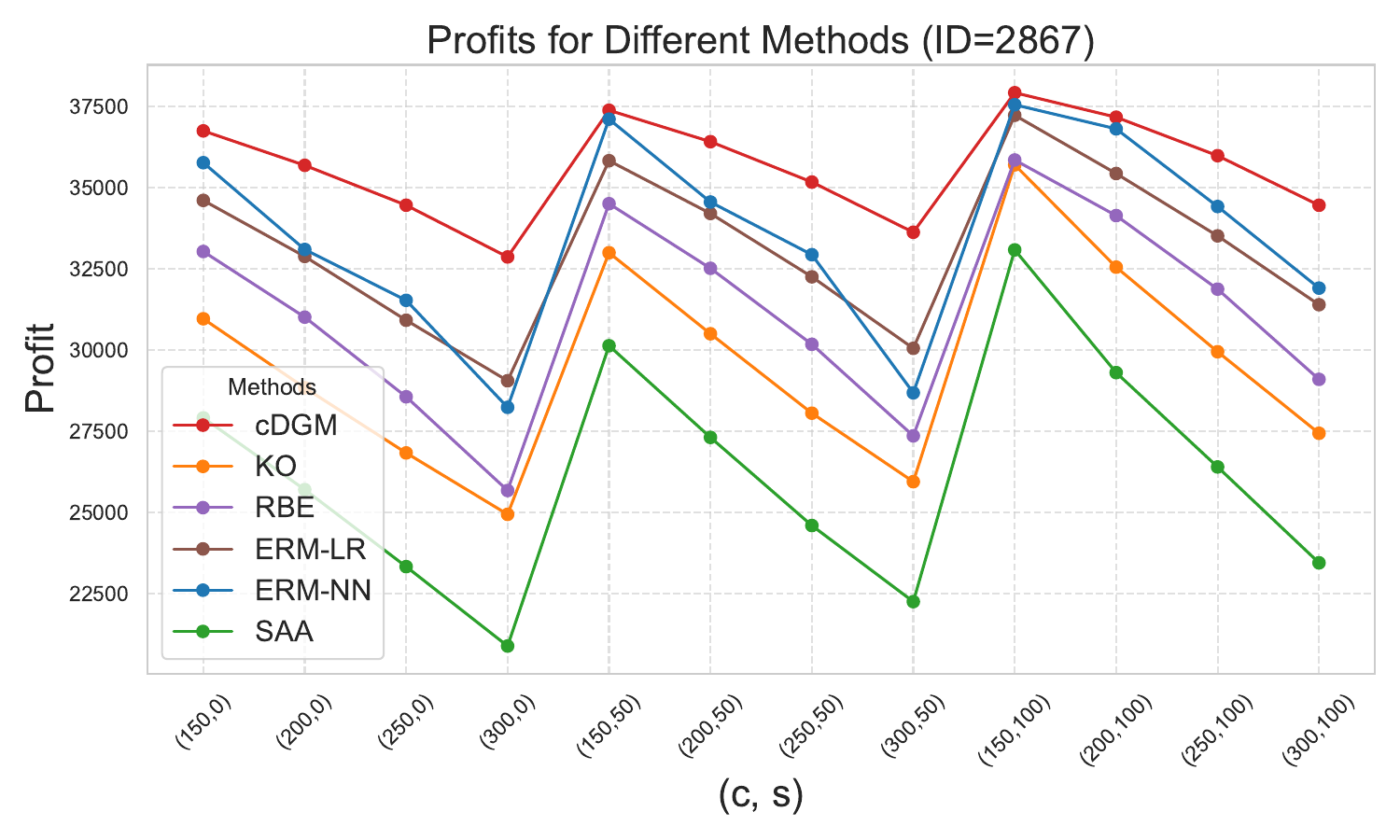}
    \end{minipage}
    \caption{Results for Kaggle Demand Dataset}
    \label{fig:real}
\end{figure}

To assess performance, we use data from the first 120 weeks for training and the remaining weeks for testing. To explore different cost scenarios, we vary \(s\) in $\{0,50,100\}$ and  \(c\) in $\{150,200,250,300\}$. Figure \ref{fig:real} shows average profits for nine meals across methods and parameter settings. In each subfigure, the x-axis represents different combinations of parameter $(c,s)$ and the y-axis represents the profit. The methods compared are the same used in Section \ref{sec:6.2.1}. We refer to Appendix B for experimental details of our methods along with the details of baseline methods that need clarification. 
The cDGM method consistently achieves the highest profits across most configurations, effectively capturing demand variability. While ERM-NN and RBE sometimes perform comparably (e.g., for meals 1982, 2581), RBE performs poorly for some meals, such as 2704. SAA generally yields the lowest profits, with KO often ranking second-lowest. These results affirm the effectiveness of our method for real-world demand forecasting.


\section{Conclusion} \label{sec:8}
In this paper, we address the feature-based newsvendor and pricing problem in a novel manner. By leveraging conditional deep generative models, we successfully tackle the complexities associated with decision-dependent uncertainty and feature-based demand distribution. Specifically, we apply cDGMs to solve both the feature-based newsvendor problem under arbitrary prices \eqref{eq:oo1} and the feature-based newsvendor pricing problem \eqref{eq:oo2}. Our proposed generative approach offers a data-driven solution to optimize decision-making without requiring explicit structural assumptions about the demand model. 

We provide theoretical guarantees, demonstrating the consistency of our profit estimation and the convergence of our decisions to optimal solutions. Extensive simulation studies, covering various scenarios—including one involving textual features—as well as a real-world case study, further confirmed the effectiveness of our approach.

Our method represents an important advancement in inventory management and pricing optimization, particularly in today’s data-rich landscape. Many modern applications, such as e-commerce platforms, social media analytics, and IoT-based systems, make extensive data collection feasible. This abundance of data, which may include multimodal sources such as text and images, presents new opportunities to enhance decision-making. Our proposed cDGM-based method is well-suited to leverage this diversity, as its deep learning foundation effectively integrates and processes complex, high-dimensional data.

It is worth noting that, while cDGMs may require more tuning than simpler models or standard neural networks, a key practical advantage of our approach is that only a single model needs to be tuned and trained for all scenarios. This model learns the conditional distribution \( D|(\bX=\bx, P=p) \) across all \( (\bx, p) \in \mathcal{X} \times \mathcal{P} \), eliminating the need to retrain or adjust the model for different configurations, as demonstrated by our experiments in Sections \ref{sec:6} and \ref{sec:7}. Furthermore, we did not perform extensive tuning in our experiments, nor did we implement advanced cDGM variants despite recent rapid AI advancements. Instead, we used a straightforward version with fully connected layers, which nonetheless achieved outstanding performance relative to other baselines.

Beyond the specific newsvendor and pricing problems addressed in this work, our approach has the potential to be extended to various other problems. We end this paper by discussing extensions of our approach: feature-dependent cost structures, multidimensional pricing and inventory decisions, and broader conditional stochastic optimization problems. These extensions are not exhaustive but indicate important directions for future research.

\subsection{Extensions and Future Work}

\subsubsection{Feature-Dependent Cost Structures}
The cost parameters $c$ and $s$ may vary depending on market conditions, product characteristics, or other features. Let \( c(\mathbf{x}) \) and \( s(\mathbf{x}) \) be two known functions that model how the purchase cost and salvage value depend on the feature vector \( \mathbf{x} \). The profit function is then expressed as:
\begin{align}
    \tilde{\Pi}(\mathbf{x},d, p, q) = (p - c(\mathbf{x})) d - (c(\mathbf{x}) - s(\mathbf{x})) (q - d)^+ - (p - c(\mathbf{x})) (d - q)^+.  \notag
\end{align}
The conditional expected profit function (\ref{eq:pipqx}) for a sales period with feature vector \( \mathbf{x} \) and joint decision \( (p, q) \) is then given by:
\begin{align}
    \pi(\mathbf{x},p, q) = \int_{-\infty}^{\infty} \tilde{\Pi}(t, p, q, \mathbf{x}) f_{D|\mathbf{X},P} (t|\mathbf{x}, p)  dt. \notag
\end{align}
We note that the cost structure does not impact the learning process of our cDGMs. Therefore, our proposed methods can solve this problem with minimal modifications, requiring only adjustments to the assigned quantile value in Algorithm \ref{algo:0} and Algorithm \ref{algo:1}.

\subsubsection{Multidimensional Price, Demand, and Inventory} \label{sec:8.1.2}
In many real-world situations, retailers must simultaneously manage multiple products or categories. The formulation can be extended to represent such multidimensional decision-making. Let \( \mathbf{p} \in \mathbb{R}^K \), \( \mathbf{d} \in \mathbb{R}^K \), \( \mathbf{q} \in \mathbb{R}^K \), $\mathbf{c}\in \mathbb{R}^K$, and $\mathbf{s}\in \mathbb{R}^K$ represent the real-valued price vector, demand vector, inventory vector, purchase cost vector, and salvage value vector, respectively, for \( K \) different products. The demand of each product is not only affected by its own price but also by the price of other products, which is also modeled by conditional distribution. The profit function for this multidimensional scenario is defined as:
\begin{align}
    \Pi(\mathbf{d}, \mathbf{p}, \mathbf{q}) = (\mathbf{p} - \mathbf{c}) \cdot \mathbf{d} - (\mathbf{c} - \mathbf{s}) \cdot (\mathbf{q} - \mathbf{d})^+ - (\mathbf{p} - \mathbf{c}) \cdot (\mathbf{d} - \mathbf{q})^+. \notag
\end{align}
The corresponding conditional expected profit function for a given feature vector $\bx$ and joint decision $(\mathbf{p},\mathbf{q})$ is:
\begin{align} \label{eq:multi1}
\pi(\mathbf{x},\mathbf{p}, \mathbf{q}) = \int_{-\infty}^{\infty} \Pi(\mathbf{t}, \mathbf{p}, \mathbf{q}) f_{\mathbf{D}|\mathbf{X},\mathbf{P}}(\mathbf{t}|\mathbf{x}, \mathbf{p}) \, d\mathbf{t}, 
\end{align}
where $f_{\mathbf{D}|\mathbf{X},\mathbf{P}}(\cdot|\mathbf{x}, \mathbf{p})$ is the conditional density. The optimal pricing and inventory decisions for each feature vector \( \mathbf{x} \) are defined by:
\begin{align} \label{eq:multi2}
    (\mathbf{p}_{\operatorname{opt}}(\mathbf{x}), \mathbf{q}_{\operatorname{opt}}(\mathbf{x})) = \underset{(\mathbf{p}, \mathbf{q}) \in \mathbb{R}^K_+ \times \mathbb{R}^K_+}{\operatorname{argmax}} \ \pi(\mathbf{p}, \mathbf{q}, \mathbf{x}). 
\end{align}
Our cDGM-based approach can be extended to learn the multidimensional conditional distribution $\mathbf{D}|\mathbf{X},\mathbf{P}$, facilitating the estimation of the profit function in \eqref{eq:multi1}. However, as the number of products $K$ increases, the optimization problem \eqref{eq:multi2} becomes increasingly complex and computationally intractable. Future research could focus on developing efficient optimization techniques to address these challenges.

\subsubsection{Conditional Stochastic Optimization}
We consider a general conditional stochastic optimization problem as introduced by \citep{bertsimas2020predictive}:
\begin{align}
    \min_{z \in \mathcal{Z}} \mathbb{E}\left[c(z ; Y) | X = x, Z = z\right], \notag
\end{align}
where \( X \in \mathcal{X} \) represents contextual features, \( Z \in \mathcal{Z} \) is the decision variable, \( Y \in \mathcal{Y} \) is a random variable whose uncertainty can be influenced by setting \( Z = z \), and \( c(z; Y) \) denotes the cost function that we aim to minimize. By training a cDGM to learn the distribution $Y|X,Z$ from the historic observations $S_n=\{(x_i,y_i,z_i)\}_{i=1}^n$, we obtain a generator \( \hat{G}: \mathcal{X} \times \mathcal{Z} \times \mathbb{R}^r \rightarrow \mathcal{Y} \), defined as \( (x, z, \eta) \mapsto y = \hat{G}(x, z, \eta) \), that approximates the conditional distribution $Y|(X=x,Z=z)$. This allows us to generate multiple samples \( \{\hat{G}(x, z, \eta^{(m)})\}_{m=1}^M \), where \( \eta^{(m)} \sim \mathcal{N}(\mathbf{0}, \mathbf{I}_r) \), providing an estimate of \( \mathbb{E}\left[c(z ; Y) | X = x, Z = z\right] \) as follows:
\begin{align} \label{eq:cso1}
    \hat{\mathbb{E}}\left[c(z ; Y) | X = x, Z = z\right] = \frac{1}{M} \sum_{m=1}^M c(z; \hat{G}(x, z, \eta^{(m)})).
\end{align}
The next step is to optimize based on the estimator in equation \eqref{eq:cso1}:
\begin{align} 
    \hat{z}_{\operatorname{cDGM}}(x) = \arg\min _{z\in\mathcal{Z}} \hat{\mathbb{E}}\left[c(z ; Y) | X = x, Z = z\right]. \notag
\end{align}
While this optimization may present challenges in different problem contexts, this paper has demonstrated the effectiveness of our approach specifically for the newsvendor pricing problem, as validated by both experimental results and theoretical analysis. This success underscores the potential of our method to be applied to other conditional stochastic optimization problems, providing a new pathway for tackling such problems within the field of management science and operations research (MS/OR).

\ACKNOWLEDGMENT{}


\bibliographystyle{informs2014} 
\bibliography{reference} 

\begin{thebibliography}{48}
\providecommand{\natexlab}[1]{#1}
\providecommand{\url}[1]{\texttt{#1}}
\providecommand{\urlprefix}{URL }

\bibitem[{Achiam et~al.(2023)Achiam, Adler, Agarwal, Ahmad, Akkaya, Aleman, Almeida, Altenschmidt, Altman, Anadkat et~al.}]{achiam2023gpt}
Achiam J, Adler S, Agarwal S, Ahmad L, Akkaya I, Aleman FL, Almeida D, Altenschmidt J, Altman S, Anadkat S, et~al. (2023) {GPT}-4 technical report. \emph{arXiv preprint arXiv:2303.08774} .

\bibitem[{Ban \protect\BIBand{} Rudin(2019)}]{ban2019big}
Ban GY, Rudin C (2019) The big data newsvendor: Practical insights from machine learning. \emph{Operations Research} 67(1):90--108.

\bibitem[{Bertsimas \protect\BIBand{} Kallus(2020)}]{bertsimas2020predictive}
Bertsimas D, Kallus N (2020) From predictive to prescriptive analytics. \emph{Management Science} 66(3):1025--1044.

\bibitem[{Beutel \protect\BIBand{} Minner(2012)}]{beutel2012safety}
Beutel AL, Minner S (2012) Safety stock planning under causal demand forecasting. \emph{International Journal of Production Economics} 140(2):637--645.

\bibitem[{Chae et~al.(2023)Chae, Kim, Kim, \protect\BIBand{} Lin}]{chae2023likelihood}
Chae M, Kim D, Kim Y, Lin L (2023) A likelihood approach to nonparametric estimation of a singular distribution using deep generative models. \emph{Journal of Machine Learning Research} 24(77):1--42.

\bibitem[{Chang et~al.(2024)Chang, Ding, Jiao, Li, \protect\BIBand{} Yang}]{chang2024deep}
Chang J, Ding Z, Jiao Y, Li R, Yang JZ (2024) Deep conditional generative learning: Model and error analysis. \emph{arXiv preprint arXiv:2402.01460} .

\bibitem[{Chen et~al.(2019)Chen, Chao, \protect\BIBand{} Ahn}]{chen2019coordinating}
Chen B, Chao X, Ahn HS (2019) Coordinating pricing and inventory replenishment with nonparametric demand learning. \emph{Operations Research} 67(4):1035--1052.

\bibitem[{Ch{\'e}rief-Abdellatif(2020)}]{cherief2020convergence}
Ch{\'e}rief-Abdellatif BE (2020) Convergence rates of variational inference in sparse deep learning. \emph{International Conference on Machine Learning}, 1831--1842 (PMLR).

\bibitem[{Dahal et~al.(2022)Dahal, Havrilla, Chen, Zhao, \protect\BIBand{} Liao}]{dahal2022deep}
Dahal B, Havrilla A, Chen M, Zhao T, Liao W (2022) On deep generative models for approximation and estimation of distributions on manifolds. \emph{Advances in Neural Information Processing Systems} 35:10615--10628.

\bibitem[{DeYong(2020)}]{deyong2020price}
DeYong GD (2020) The price-setting newsvendor: {R}eview and extensions. \emph{International Journal of Production Research} 58(6):1776--1804.

\bibitem[{Ding et~al.(2024)Ding, Huh, \protect\BIBand{} Rong}]{ding2024feature}
Ding J, Huh WT, Rong Y (2024) Feature-based inventory control with censored demand. \emph{Manufacturing \& Service Operations Management} 26(3):1157--1172.

\bibitem[{Dinh et~al.(2014)Dinh, Krueger, \protect\BIBand{} Bengio}]{dinh2014nice}
Dinh L, Krueger D, Bengio Y (2014) {NICE}: Non-linear independent components estimation. \emph{arXiv preprint arXiv:1410.8516} .

\bibitem[{Fu et~al.(2024)Fu, Li, \protect\BIBand{} Zhang}]{fu2024distributionally}
Fu M, Li X, Zhang L (2024) Distributionally robust newsvendor under stochastic dominance with a feature-based application. \emph{Manufacturing \& Service Operations Management} 26(5):1962--1977.

\bibitem[{Goodfellow et~al.(2020)Goodfellow, Pouget-Abadie, Mirza, Xu, Warde-Farley, Ozair, Courville, \protect\BIBand{} Bengio}]{goodfellow2020generative}
Goodfellow I, Pouget-Abadie J, Mirza M, Xu B, Warde-Farley D, Ozair S, Courville A, Bengio Y (2020) Generative adversarial networks. \emph{Communications of the ACM} 63(11):139--144.

\bibitem[{Han et~al.(2023)Han, Hu, \protect\BIBand{} Shen}]{han2023deep}
Han J, Hu M, Shen G (2023) Deep neural newsvendor. \emph{arXiv preprint arXiv:2309.13830} .

\bibitem[{Han et~al.(2022)Han, Zheng, \protect\BIBand{} Zhou}]{han2022card}
Han X, Zheng H, Zhou M (2022) {CARD}: Classification and regression diffusion models. \emph{Advances in Neural Information Processing Systems} 35:18100--18115.

\bibitem[{Harsha et~al.(2021)Harsha, Natarajan, \protect\BIBand{} Subramanian}]{harsha2021prescriptive}
Harsha P, Natarajan R, Subramanian D (2021) A prescriptive machine-learning framework to the price-setting newsvendor problem. \emph{Informs Journal on Optimization} 3(3):227--253.

\bibitem[{Hastie et~al.(2009)Hastie, Tibshirani, Friedman, \protect\BIBand{} Friedman}]{hastie2009elements}
Hastie T, Tibshirani R, Friedman JH, Friedman JH (2009) \emph{The {E}lements of {S}tatistical {L}earning: Data Mining, Inference, and Prediction} (Springer), 2 edition.

\bibitem[{Ho et~al.(2020)Ho, Jain, \protect\BIBand{} Abbeel}]{ho2020denoising}
Ho J, Jain A, Abbeel P (2020) Denoising diffusion probabilistic models. \emph{Advances in Neural Information Processing Systems} 33:6840--6851.

\bibitem[{Huang et~al.(2013)Huang, Leng, \protect\BIBand{} Parlar}]{huang2013demand}
Huang J, Leng M, Parlar M (2013) Demand functions in decision modeling: A comprehensive survey and research directions. \emph{Decision Sciences} 44(3):557--609.

\bibitem[{Jiao et~al.(2023)Jiao, Shen, Lin, \protect\BIBand{} Huang}]{jiao2023deep}
Jiao Y, Shen G, Lin Y, Huang J (2023) Deep nonparametric regression on approximate manifolds: Nonasymptotic error bounds with polynomial prefactors. \emph{The Annals of Statistics} 51(2):691--716.

\bibitem[{Kincaid \protect\BIBand{} Darling(1963)}]{kincaid1963inventory}
Kincaid W, Darling D (1963) An inventory pricing problem. \emph{Journal of Mathematical Analysis and Applications} 7(2):183--208.

\bibitem[{Kingma(2013)}]{kingma2013auto}
Kingma DP (2013) Auto-encoding variational bayes. \emph{arXiv preprint arXiv:1312.6114} .

\bibitem[{Li et~al.(2022)Li, Lu, Wang, \protect\BIBand{} Dou}]{li2022generative}
Li Y, Lu X, Wang Y, Dou D (2022) Generative time series forecasting with diffusion, denoise, and disentanglement. \emph{Advances in Neural Information Processing Systems} 35:23009--23022.

\bibitem[{Liu et~al.(2021)Liu, Zhou, Jiao, \protect\BIBand{} Huang}]{liu2021wasserstein}
Liu S, Zhou X, Jiao Y, Huang J (2021) Wasserstein generative learning of conditional distribution. \emph{arXiv preprint arXiv:2112.10039} .

\bibitem[{Liu \protect\BIBand{} Zhang(2023)}]{liu2023solving}
Liu W, Zhang Z (2023) Solving data-driven newsvendor pricing problems with decision-dependent effect. \emph{arXiv preprint arXiv:2304.13924} .

\bibitem[{Liyanage \protect\BIBand{} Shanthikumar(2005)}]{liyanage2005practical}
Liyanage LH, Shanthikumar JG (2005) A practical inventory control policy using operational statistics. \emph{Operations Research Letters} 33(4):341--348.

\bibitem[{Mills(1959)}]{mills1959uncertainty}
Mills ES (1959) Uncertainty and price theory. \emph{The Quarterly Journal of Economics} 73(1):116--130.

\bibitem[{Oko et~al.(2023)Oko, Akiyama, \protect\BIBand{} Suzuki}]{oko2023diffusion}
Oko K, Akiyama S, Suzuki T (2023) Diffusion models are minimax optimal distribution estimators. \emph{International Conference on Machine Learning}, 26517--26582 (PMLR).

\bibitem[{Oroojlooyjadid et~al.(2020)Oroojlooyjadid, Snyder, \protect\BIBand{} Tak{\'a}{\v{c}}}]{oroojlooyjadid2020applying}
Oroojlooyjadid A, Snyder LV, Tak{\'a}{\v{c}} M (2020) Applying deep learning to the newsvendor problem. \emph{IISE Transactions} 52(4):444--463.

\bibitem[{Ouyang et~al.(2022)Ouyang, Wu, Jiang, Almeida, Wainwright, Mishkin, Zhang, Agarwal, Slama, Ray et~al.}]{ouyang2022training}
Ouyang L, Wu J, Jiang X, Almeida D, Wainwright C, Mishkin P, Zhang C, Agarwal S, Slama K, Ray A, et~al. (2022) Training language models to follow instructions with human feedback. \emph{Advances in Neural Information Processing Systems} 35:27730--27744.

\bibitem[{Petruzzi \protect\BIBand{} Dada(1999)}]{petruzzi1999pricing}
Petruzzi NC, Dada M (1999) Pricing and the newsvendor problem: A review with extensions. \emph{Operations Research} 47(2):183--194.

\bibitem[{Qi et~al.(2023)Qi, Shi, Qi, Ma, Yuan, Wu, \protect\BIBand{} Shen}]{qi2023practical}
Qi M, Shi Y, Qi Y, Ma C, Yuan R, Wu D, Shen ZJ (2023) A practical end-to-end inventory management model with deep learning. \emph{Management Science} 69(2):759--773.

\bibitem[{Qin et~al.(2022)Qin, Simchi-Levi, \protect\BIBand{} Wang}]{qin2022data}
Qin H, Simchi-Levi D, Wang L (2022) Data-driven approximation schemes for joint pricing and inventory control models. \emph{Management Science} 68(9):6591--6609.

\bibitem[{Qin et~al.(2011)Qin, Wang, Vakharia, Chen, \protect\BIBand{} Seref}]{qin2011newsvendor}
Qin Y, Wang R, Vakharia AJ, Chen Y, Seref MM (2011) The newsvendor problem: Review and directions for future research. \emph{European Journal of Operational Research} 213(2):361--374.

\bibitem[{Rasul et~al.(2021)Rasul, Seward, Schuster, \protect\BIBand{} Vollgraf}]{rasul2021autoregressive}
Rasul K, Seward C, Schuster I, Vollgraf R (2021) Autoregressive denoising diffusion models for multivariate probabilistic time series forecasting. \emph{International Conference on Machine Learning}, 8857--8868 (PMLR).

\bibitem[{Rombach et~al.(2022)Rombach, Blattmann, Lorenz, Esser, \protect\BIBand{} Ommer}]{rombach2022high}
Rombach R, Blattmann A, Lorenz D, Esser P, Ommer B (2022) High-resolution image synthesis with latent diffusion models. \emph{Proceedings of the IEEE/CVF Conference on Computer Vision and Pattern Recognition}, 10684--10695.

\bibitem[{Sachs \protect\BIBand{} Minner(2014)}]{sachs2014data}
Sachs AL, Minner S (2014) The data-driven newsvendor with censored demand observations. \emph{International Journal of Production Economics} 149:28--36.

\bibitem[{Schmidt-Hieber(2020)}]{schmidt2020nonparametric}
Schmidt-Hieber J (2020) Nonparametric regression using deep neural networks with {ReLU} activation function. \emph{The Annals of Statistics} 48(4):1875--1897.

\bibitem[{See \protect\BIBand{} Sim(2010)}]{see2010robust}
See CT, Sim M (2010) Robust approximation to multiperiod inventory management. \emph{Operations Research} 58(3):583--594.

\bibitem[{Shapiro(2003)}]{shapiro2003monte}
Shapiro A (2003) {Monte Carlo} sampling methods. \emph{Handbooks in Operations Research and Management Science} 10:353--425.

\bibitem[{Tomczak(2022)}]{Tomczak202DeepGenerative}
Tomczak JM (2022) \emph{Deep Generative Modeling} (Springer), ISBN 978-3-030-93157-5.

\bibitem[{Whitin(1955)}]{whitin1955inventory}
Whitin T (1955) Inventory control and price theory. \emph{Management Science} 2(1):61--68.

\bibitem[{Young(1978)}]{young1978price}
Young L (1978) Price, inventory and the structure of uncertain demand. \emph{New Zealand Operations Research} 6(2):157--177.

\bibitem[{Zhang et~al.(2024)Zhang, Yang, \protect\BIBand{} Gao}]{zhang2024optimal}
Zhang L, Yang J, Gao R (2024) Optimal robust policy for feature-based newsvendor. \emph{Management Science} 70(4):2315--2329.

\bibitem[{Zhao et~al.(2024)Zhao, Zhou, \protect\BIBand{} Wang}]{zhao2024private}
Zhao T, Zhou WX, Wang L (2024) Private optimal inventory policy learning for feature-based newsvendor with unknown demand. \emph{Management Science} .

\bibitem[{Zheng et~al.(2024)Zheng, Li, Jiang, \protect\BIBand{} Peng}]{zheng2024dual}
Zheng Y, Li Z, Jiang P, Peng Y (2024) Dual-agent deep reinforcement learning for dynamic pricing and replenishment. \emph{arXiv preprint arXiv:2410.21109} .

\bibitem[{Zhou et~al.(2023)Zhou, Jiao, Liu, \protect\BIBand{} Huang}]{zhou2023deep}
Zhou X, Jiao Y, Liu J, Huang J (2023) A deep generative approach to conditional sampling. \emph{Journal of the American Statistical Association} 118(543):1837--1848.

\end{thebibliography}


\end{document}